\RequirePackage{fix-cm}
\documentclass[twocolumn]{svjour3}
\smartqed
\usepackage{natbib}
\usepackage{graphicx}
\usepackage{booktabs}
\usepackage{scalefnt}
\usepackage{amssymb}
\usepackage{adjustbox}
\usepackage{xspace}
\usepackage{url}
\usepackage{subfigure}
\usepackage{soul, color}
\usepackage{graphicx}
\usepackage{amsmath}
\usepackage{xcolor}

\newcommand{\name}{{\it QuaterNet}\xspace}

\newcommand{\citeconf}[2]{{(\citeauthor{#1}, #2 \citeyear{#1})}}

\begin{document}

\title{Modeling Human Motion with Quaternion-based Neural Networks}

\author{Dario Pavllo \and
        Christoph Feichtenhofer \and
        Michael Auli \and
        David Grangier
}

\institute{Dario Pavllo \at
              ETH Zurich \\
              \textit{Majority of work done during an internship at Facebook AI Research.} \\
           \and
           Christoph Feichtenhofer \and Michael Auli \at
           Facebook AI Research \\
           \and
           David Grangier \at
           Google Brain \\
           \textit{Work done while at Facebook AI Research.}
}

\date{}

\maketitle

\begin{abstract}
\sloppy
Previous work on predicting or generating 3D human pose sequences regresses either joint rotations or joint positions. The former strategy is prone to error accumulation along the kinematic chain, as well as discontinuities when using Euler angles or exponential maps as parameterizations. The latter requires re-projection onto skeleton constraints to avoid bone stretching and invalid configurations. This work addresses both limitations. \name represents rotations with quaternions and our loss function performs forward kinematics on a skeleton to penalize absolute position errors instead of angle errors. We investigate both recurrent and convolutional architectures and evaluate on short-term prediction and long-term generation. For the latter, our approach is qualitatively judged as realistic as recent neural strategies from the graphics literature. Our experiments compare quaternions to Euler angles as well as exponential maps and show that only a very short context is required to make reliable future predictions. Finally, we show that the standard evaluation protocol for Human3.6M produces high variance results and we propose a simple solution.

\end{abstract}

%----------------------------------------------------
\section{Introduction}
\label{sec:intro}
\sloppy

Modeling human motion is useful for many applications, including human action recognition \citep{du:action:2015}, action detection \citep{gu:ava:2018}, or action anticipation \citep{kitani:activity:2012}. Forecasting human motion trajectories is essential for applications in robotics \citep{koppula:anticipating:2016} or computer graphics \citep{holden:deeplearning:2016}.
Deep learning-based approaches have been successful in other pattern recognition tasks \citep{krizhevsky:imagenet:2012,hinton:speech:2012,bahdanau:attention:2015}, and they have also been studied for the prediction of sequences of 3D-skeleton joint positions (i.e.~3D human pose), both for short-term \citep{fragkiadaki:recurent:2015,martinez:recurrent:2017} and long-term modeling \citep{holden:deeplearning:2016,holden:phase:2017}. 

Human motion is a stochastic sequential process with a high level of intrinsic uncertainty. Given an observed sequence of poses, a rich set of future pose sequences are likely, depending on factors such as physics or the conscious intentions of a person. 
Therefore, predictions far in the future are unlikely to match a reference recording, even with an excellent model.  
Consequently, the literature often distinguishes between short-term and long-term prediction tasks. Short-term tasks are often referred to as \emph{prediction} tasks and can be assessed quantitatively by comparing the model prediction to a reference recording through a distance metric. Long-term tasks are often referred to as \emph{generation} tasks and are harder to assess quantitatively. For these cases, the prediction quality can be evaluated by human evaluation studies.

This work addresses both short-term and long-term tasks through a unified approach, with the goal of competing with state-of-the-art methods in the computer vision literature for short-term prediction, as well as to compete with the state-of-the-art in the computer graphics literature for long-term generation. With that objective in mind, we identify the limitations of current approaches and address them. Our contributions are threefold. First, we propose a methodology for employing a quaternion-based pose representation in recurrent and convolutional neural networks. Other parameterizations, such as Euler angles, suffer from discontinuities and singularities, which can lead to exploding gradients and difficulty in training the model. Previous work \citep{taylor:binarylatent:2006,martinez:recurrent:2017} tried to mitigate these issues by switching to \emph{exponential maps} (also referred to as \emph{axis-angle representation}), which makes them less likely to exhibit these issues but does not solve them entirely \citep{grassia:rotations:1998}. Second, we propose a differentiable loss function which conducts forward kinematics on a parameterized skeleton, and combines the advantages of joint orientation prediction with those of a position-based loss. Finally, we point out a flaw in the standard evaluation protocol of the Human3.6M dataset which causes the results to have high variance and we propose a simple adjustment to mitigate this issue.

We conduct experiments on short-term prediction and long-term generation, evaluating the former on the Human3.6M benchmark \citep{ionescu:human36:2014} and the latter on the locomotion dataset from~\cite{holden:deeplearning:2016}.
Short-term performance is slightly outperformed by very recent work on adversarial training \citep{gui:adversarial:2018}. Adversarial training and quaternion-based parameterization are however orthogonal aspects in motion modeling. Their combination is beyond the scope of this study and is surely an interesting path to future improvement.
Long-term generation quality matches the quality of recent work from the computer graphics literature, while allowing on-line generation, and better control over the timings and trajectory constraints imposed by the artist.

This article extends~\citet{pavllo:quaternet_bmvc:2018} as follows:
\begin{itemize}
    \item We introduce a version of \name based on a convolutional neural network and compare to the original recurrent neural network approach.
    \item We empirically compare alternatives to quaternions and contrast them to Euler angles as well as exponential maps.
    \item We ablate the amount of temporal context that is required to make reliable future predictions and find that a relatively short context results in as good performance as longer context.
    \item We address a flaw in the standard evaluation methodology and propose a variant that yields more stable results.
\end{itemize}

The remainder of the paper examines related work (Section~\ref{sec:related}),
describes our \name method (Section~\ref{sec:method}) and presents our experiments (Section~\ref{sec:exp}). Finally, we draw some conclusions and delineate potential future work (Section~\ref{sec:ccl}). 
We also release our code and pre-trained models publicly at\\ 
\url{https://github.com/facebookresearch/QuaterNet}

\section{Related work}
\label{sec:related}

The modeling of human motion relies on data from motion capture. This technology acquires sequences of 3-dimensional joint positions at high frame rate (120~Hz~--~1~kHz) and enables a wide range of applications, such as performance animation in movies and video games, and motion generation. In that context, the task of generating human motion sequences has been addressed with different strategies ranging from purely concatenation-based approaches \citep{arikan:motion:2003}, concatenate-and-blend \citep{treuille:nearopt:2007}, to hidden Markov models \citep{tanco:hmm:2000}, switching linear dynamic systems \citep{pavlovic:slds:2000}, restricted Boltzmann machines \citep{taylor:binarylatent:2006}, Gaussian processes \citep{wang:gp:2008}, and random forests \citep{lehrmann:hmm:14}.

Recently, Recurrent Neural Networks (RNN) have been applied to short \citep{fragkiadaki:recurent:2015,martinez:recurrent:2017} and long-term prediction \citep{zhou:lstm:2018}. Convolutional networks \citep{holden:deeplearning:2016,li:convnet:2018} and feed-forward networks \citep{holden:phase:2017} have been successfully applied to long-term generation of locomotion. Early work took great care in choosing a model expressing the inter-dependence between joints \citep{jain:srnn:2016}, while recent work favors universal approximators \citep{martinez:recurrent:2017,butepage:2017,holden:deeplearning:2016,holden:phase:2017}. 
Beside choosing the neural architecture, framing the pose prediction task is equally important. Specifically, defining input and output variables, their representation as well as the loss function used for training are particularly impactful, as we show in our experiments.
Equally important are the control variables conditioning motion generation. Long-term generation is an highly under-specified task with high uncertainty. In practice, animators for movies and games are interested in motion generators that can be conditioned from high level controls like trajectories and velocities \citep{holden:phase:2017}, style \citep{li:synthesis:2018} or action classes \citep{kiasari:actiongan:2018}. Game development tools typically rely on classical move trees \citep{menache:mocap:1999}, which allows for a wide range of controls and excellent run-time efficiency. These advantages comes with a high development effort to deal with all possible action transitions. The development cost of move trees makes learning-based approach an attractive area of research.

As for quaternions in neural networks, \cite{gaudet:deep:2018} propose a hyper-complex extension of complex-valued convolutional neural networks, and \cite{kumar:ijies:2017} present a variation of resilient backpropagation in quaternionic domain. The motivation of these works is different than ours. Their work shows that quaternionic domain latent variables can encode long term-dependencies with fewer learned parameters than real-valued models. In our case, we rely on quaternions for the representation of rotations along a kinematic chain, a classical formulation in computer graphics \citep{mccarthy1990introduction}, see Section~\ref{sec:quaternion_loss}.

\subsection{Joint rotations versus positions}
\label{sec:rotations_vs_positions}

Human motion is represented as a sequence of human poses. Each pose can be described through body joint positions, or through 3D-joint rotations which are then integrated via forward kinematics. For motion prediction, one can consider predicting either rotations or positions with alternative benefits and trade-offs. Depending on the application, a particular representation may be required: for instance, in video games and movies it is typical to animate a skinned mesh using joint rotations.
 
The prediction of rotations allows using a parameterized skeleton \citep{pavlovic:slds:2000,taylor:binarylatent:2006,fragkiadaki:recurent:2015}. Skeleton constraints avoid prediction errors such as non-constant bone lengths or motions outside an articulation range. However, rotation prediction is often paired with a loss that averages errors over joints which gives each joint the same weight. This ignores that the prediction errors of different joints have varying impact on the body, e.g. joints between the trunk and the limbs typically impact the pose more than joints at the end of limbs, with the root joint being the extreme case. This type of loss can therefore yield a model with spurious large errors on important joints, which severely impact generation from a qualitative perspective.

The prediction of joint positions minimizes the averaged position errors over 3D points, and as such does not suffer from this problem. However, this strategy does not benefit from the parameterized skeleton constraints and needs its prediction to be reprojected onto a valid configuration to avoid issues like bone stretching \citep{holden:deeplearning:2016,holden:phase:2017}. This step can be resource intensive and is less efficient in terms of model fitting. When minimizing the loss, model fitting ignores that the prediction will be reprojected onto the skeleton, which often increases the loss. Also, the projection step can yield discontinuities in time, as we show in Section~\ref{sec:ablations}.

Alternatively one can choose to learn a network which does not predict positions, while still minimizing position errors. This is performed by mapping the outputs of the network to positions with a differential transformation. For hand pose estimation, \citep{oberweger:pca:2015} introduces a network which outputs a latent representation of the hand that can be linearly projected to positions. This representation is obtained through Principal Component Analysis learned from the position vectors prior to training \citep{oberweger:pca:2015,cootes:asm:2000}. In that line of work, joint rotations can be mapped to positions through forward kinematics over a parameterized skeleton. This operation is differentiable and has been used to train networks for hand tracking \citep{zhou:ijcai:2016} and pose estimation from still images \citep{zhou:eccvw:2016}. Our work builds upon this strategy.

For both positions and rotations, one can consider predicting velocities (i.e. deltas w.r.t. time) instead of absolute values \citep{martinez:recurrent:2017,toyer:dicta:2017}. The density of velocities is concentrated in a smaller range of values, which helps statistical learning. However, in practice velocities tend to be unstable in long-term tasks, and generalize worse due to accumulation of errors. Noise in the training data is also problematic with velocities: invalid poses introduce large variations which can yield unstable models.

Alternatively to the direct modeling of joint rotations/positions, physics-inspired models of the human body have also been explored \citep{liu:physics:2005} but such models have been less popular for generation with the availability of larger motion capture datasets \citep{cmu:mocap,muller:hdm05:2007,ionescu:human36:2014}.

\subsection{Learning a stochastic process}

Human motion is a stochastic process with a high level of uncertainty. For a given past, there will be multiple likely sequences of future frames and uncertainty grows with duration. This makes training for long-term generation challenging since recorded frames far in the future will capture only a small fraction of the probability mass, even according to a perfect model. 

Like other stochastic processes \citep{bengio:lm:2003,oord:wavenet:2016,oord:pixelrnn:2016}, motion modeling is often addressed by training transition operators, also called auto-regressive models. At each time step, such a model predicts the next pose given the previous poses. Typically, training such a model involves supplying recorded frames to predict the next recorded target. This strategy -- called teacher forcing -- does not expose the model to its own errors and prevents it from recovering from them, a problem known as \emph{exposure bias} \citep{ranzato:mixer:2015,wiseman:bso:2016}. To mitigate this problem, previous work suggested to add noise to the network inputs during training \citep{fragkiadaki:recurent:2015,ghosh:dropout:2017}. Alternatively, \cite{martinez:recurrent:2017} forgo teacher forcing and always inputs model predictions. This strategy however can yield slow training since the loss can be very high on long sequences. 

Due to the difficulty of long-term prediction, previous work has considered decomposing this task hierarchically. For locomotion, \cite{holden:deeplearning:2016} propose to subdivide the task into three steps: define the character trajectory, annotate the trajectory with footsteps, generate pose sequences. The neural network for the last step takes trajectory and speed data as input. This strategy makes the task simpler since the network is relieved from modeling the uncertainty due to the trajectory and walk cycle drift. \cite{holden:phase:2017} consider a network which computes different sets of weights according to the phase in the walk cycle. Other work consider alternative metrics and human evaluation to deal with the uncertainty of the task \citep{gopalakrishnan:neurotemporal:2018}. 

Most research casts the problem of motion prediction of the next frame as a regression problem, without explicitly modeling uncertainty. Such models can only predicts the expectation of the next pose, which can be a problem for multi-modal data. Neural generative modeling addresses this problem, including Generative Adversarial Networks \citep{mathieu:iclr:2016,luc:iccv:2017} and Variational Auto-Encoders \citep{walker:eccv:2016}. Both GANs \citep{villegas:icml:2017,kiasari:actiongan:2018,gui:adversarial:2018,lin:dvgan:2018,wang:gan:2018} and VAEs \citep{walker:iccv:2017,butepage:anticipating:2018} have been applied to the task of human motion prediction.
A recent work, \citep{gui:adversarial:2018}, is of particular interest, as it shows strong performance by proposing two distinct discriminators learned jointly with the sequence generator. A classical discriminator tries to distinguish the model generation from real data, while a second discriminator focuses on distinguishes whether generation conditioned on a true prefix sequences produces realistic continuations.  

\subsection{Pose and video forecasting}

Forecasting is an active topic of research beyond the prediction of human pose sequences. Pixel-level prediction using human pose as an intermediate variable has been explored \citep{villegas:icml:2017,walker:iccv:2017}. Related work also includes the forecasting of locomotion trajectories \citep{kitani:eccv:2012}, human instance segmentation \citep{luc:arxiv:2018}, or future actions \citep{lan:eccv:2014}. 
Other types of conditioning have also been explored for predicting poses: for instance, \cite{shlizerman:audio:2017} explore generating skeleton pose sequences of music players from audio, \cite{chao:cvpr:2017} aim at predicting future pose sequences from static images. Also relevant is the prediction of 3D poses from images or 2D joint positions \citep{parameswaran:single:2004,radwan:monocular:2013,akhter:cvpr:2015}. The prediction of rigid object motion for robotic applications is also relevant, e.g. \cite{byravan:icra:2017} model object dynamics using a neural network that performs spatial transformations on point clouds.

\section{QuaterNet}
\label{sec:method}

This section introduces our quaternion-based neural architectures for modeling human motion. It first describes a recurrent architecture and then a convolutional version. Next, we detail our training procedure and then discuss forward kinematics as well as rotation parameterizations. Finally, we describe specifics of our short and long-term motion models. 

\subsection{Recurrent architecture}

In the original formulation of \name \citep{pavllo:quaternet_bmvc:2018}, we use an RNN to model sequences of three-dimensional poses as in \cite{fragkiadaki:recurent:2015} and \cite{martinez:recurrent:2017}. 
We have a two-layer \emph{gated recurrent unit} (GRU) network \citep{cho:gru:2014} that is an autoregressive model, i.e. at each time step, the model takes as input the previous recurrent state as well as features describing the previous pose in order to predict the next pose. 
Similar to \cite{martinez:recurrent:2017}, we selected GRU for their simplicity and efficiency. In line with the findings of \cite{chung:empirical:2014}, we found no benefit in using \emph{long short-term memory} (LSTM), which require learning extra gates. Contrary to \cite{martinez:recurrent:2017}, however, we found an empirical advantage of adding a second recurrent layer, but not a third one. The two GRU layers comprise $1,000$ hidden units each, and their initial states $\mathbf{h_0}$ are learned from the data.

\begin{figure*}
(a) Recurrent architecture for short-term prediction\\
\includegraphics[width=\textwidth]{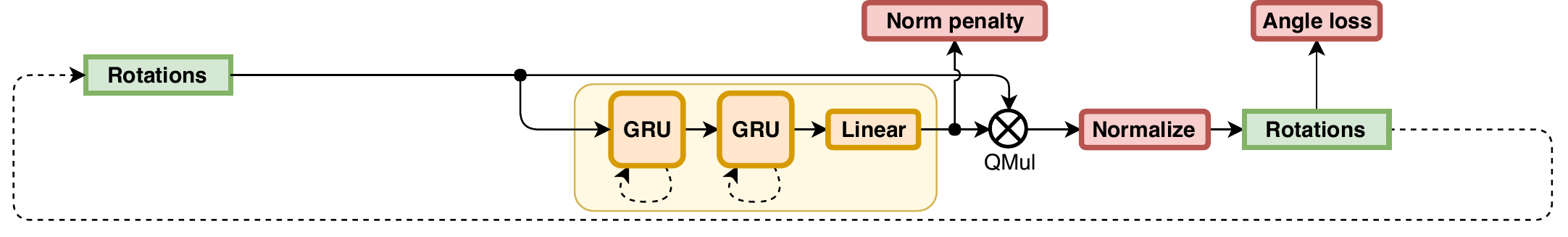}\\
(b) Recurrent architecture for long-term generation\\
\includegraphics[width=\textwidth]{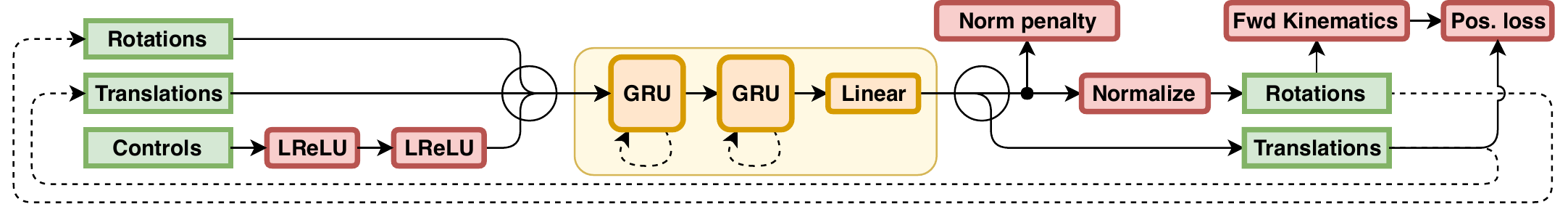}
\caption{Recurrent architecture. ``QMul'' stands for quaternion multiplication: if included, it forces the model to output velocities; if bypassed, the model emits absolute rotations. The center block (in yellow) is the recurrent backbone of the network.}
\label{fig:architecture}
\end{figure*}

Figure~\ref{fig:architecture} shows the high-level architecture of our \emph{pose network}, which we use for both short-term prediction and long-term generation. If employed for the latter purpose, the model includes additional inputs (referred to as ``Translations'' and ``Controls'' in the figure), which are used to provide artistic control. The network takes as input the rotations of all joints (encoded as unit quaternions, a choice that we motivate in Section~\ref{sec:quaternion_loss}), plus optional inputs, and is trained to predict the future states of the skeleton across $k$ time steps, given $n$ frames of initialization; $k$ and $n$ depend on the task.
\begin{figure}
\includegraphics[width=\linewidth]{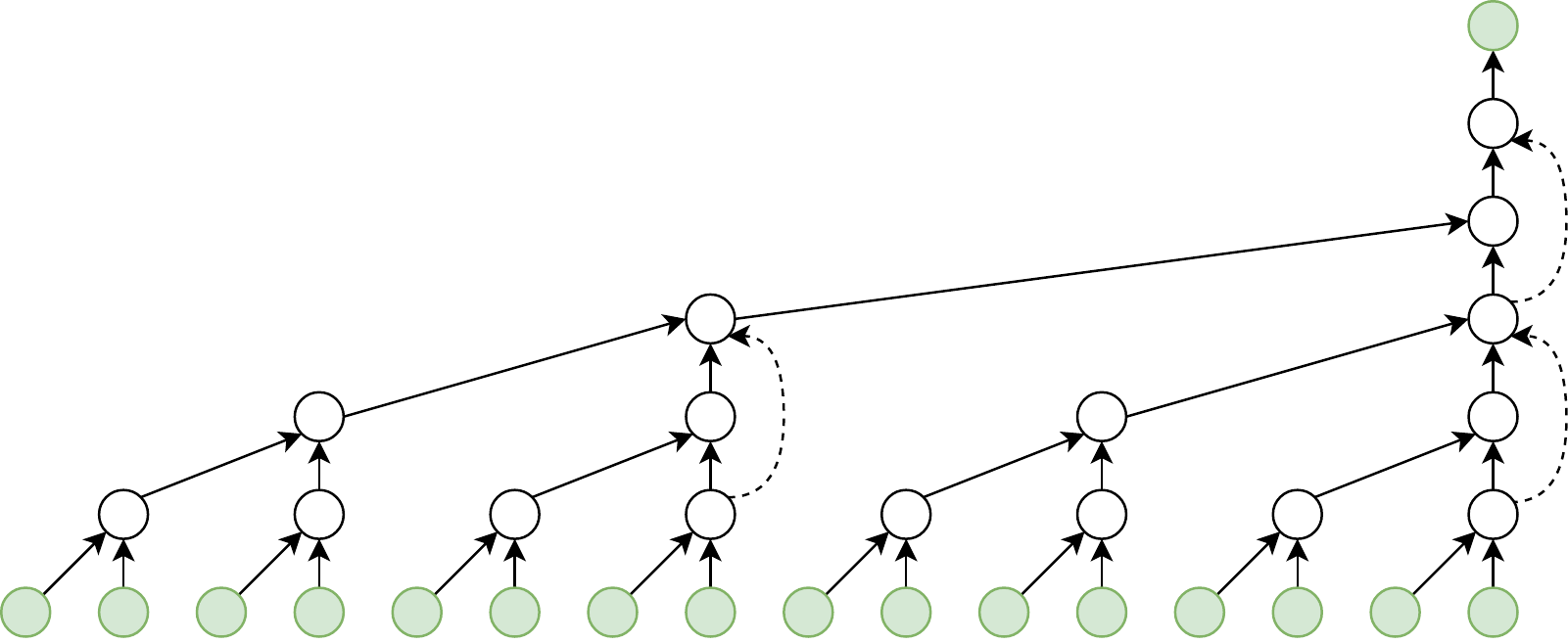}
\caption{Convolutional backbone which can replace the recurrent backbone in Figure~\ref{fig:architecture} (shaded there in yellow). Dashed lines represent skip-connections. We depict a model with a receptive field of 16 frames ($W = 1$ in the last layer).}
\label{fig:architecture_cnn}
\end{figure}

\subsection{Convolutional architecture}
A recent trend in sequence modeling consists in replacing RNNs with convolutional neural networks (CNN) for tasks that were typically tackled with the former. These include neural machine translation \citep{gehring:convs2s:2017}, language modeling~\citep{dauphin:2016:convlm}, speech processing~\citep{collobert:2016:wav2letter}, and 3D human pose estimation in video~\citep{pavllo:videopose3d:2018}, where convolutional architectures have achieved compelling results. 

Compared to RNNs, convolutional networks have a number of advantages. First, they are more efficient on modern hardware since they can be parallelized both across the batch and time/space dimensions. 
Recurrent models can only be parallelized across the batch dimension due to their dependence on previous time-steps. 
Second, training is simpler since convolutional architectures have a constant path length between the input and the output, which makes them less likely to suffer under exploding or vanishing gradients such as RNNs. 
On the other hand, RNNs are in theory able to model arbitrary length sequences with a fixed number of parameters.
However, in practice they tend to focus on local dependencies rather than long-term relationships.
In convolutional models, the receptive field can be drastically increased through \emph{dilated convolutions}, which result in the number of parameters to grow only logarithmically with respect to the receptive field.

To better understand whether convolutional architectures can be beneficial for human motion modeling, we introduce a variation of \name based on temporal convolutions and analyze it. 
Our convolutional architecture is an adaptation of its RNN-based counterpart, in which we replace the backbone (GRU and linear layers, yellow block in Figure~\ref{fig:architecture}) with a sequence of convolutional layers.

We adopt convolutions with filter width $W = 2$ and an exponentially increasing dilation factor $D = 2^k$, where $k$ is the current layer (from 1 to 5, i.e. 5 layers in total). 
This strategy ensures that the path from the input to the output forms a tree in which each input frame is read exactly once by the first layer and each output of the first layer is processed only once by the second layer and so on.
Our convolutions are \emph{causal}, i.e. they only look at past frames. 
The receptive field can be controlled precisely by varying $W$, e.g. if $W = 2$ for all layers we obtain a receptive field of 32 frames; if we set $W = 3$ in the last layer, then we get 48 frames, and so on. 
We also add skip-connections between every other layer, as these make it easier to propagate gradients through multiple layers \citep{he:deep:2016}. 
Similar to the recurrent velocity model, we multiply the output quaternions with the input in order to force the model to represent rotation deltas internally. 
All convolutions use $C = 1024$ channels, except the first and last layer, which map from and to the number of rotation parameters. 
The information flow in our convolutional architecture is depicted in Figure~\ref{fig:architecture_cnn}.

As an ablation, we tried to replace \emph{dilated} convolutions with standard dense convolutions, but this did not result in any improvements. Dilated convolutions perform consistently better, suggesting that they generalize more easily due to their sparsity.

\subsection{Training details}
For optimization, we use Adam~\citep{kingma:adam:2014} and we clip the gradient norm to $0.1$. The learning learning rate is decayed exponentially with a factor of $\alpha = 0.999$ per epoch. For efficient batching, we sample fixed length episodes from the training set, sampling uniformly across valid starting points. We define an epoch to be a random sample of size equal to the number of sequences.

To address the challenging task of generating long-term motion, the network is progressively exposed to its own predictions through a curriculum schedule known as \emph{scheduled sampling}~\citep{bengio:scheduled:2015}. We found the latter to be beneficial for improving the error and model stability, as we demonstrate in Figure~\ref{fig:schedule_vs_sampling}. At every time step, we randomly sample from a Bernoulli distribution with probability $p$ to determine whether the model should observe the ground truth or its own prediction. Initially, we set $p = 1$ (i.e. teacher forcing), and we decay it exponentially with a factor $\beta = 0.995$ per epoch.

When the recurrent architecture is exposed to its own predictions, then the derivative of the loss with respect to its output sums two terms: the first term makes the current prediction closer to the current target and the second term adjusts the current prediction to improve future predictions. 
In the convolutional architecture the gradient flows only across the first term, as in~\citet{bengio:scheduled:2015}. 
Also, we train both CNNs and RNNs without layer normalization \citep{ba:layer:2016} or batch normalization~\citep{ioffe:batch:2015} as neither led to improvements in our setting.

\subsection{Parameterization of forward kinematics}
\label{sec:quaternion_loss}

Euler angles are often used to represent joint rotations \citep{han:survey:2017}. They offer the advantage to specify an angle for each degree of freedom, so they can be easily constrained to match the degrees of freedom of real human joints. However, Euler angles also suffer from non-uniqueness ($\alpha$ and $\alpha + 2 \pi n$ represent the same angle), discontinuity in the representation space, and singularities (\emph{gimbal lock}). It can be shown that all representations in $\mathbb{R}^3$ suffer from these problems, including the popular exponential maps \citep{grassia:rotations:1998}. In contrast, quaternions -- which lie in $\mathbb{R}^4$ -- are free of discontinuities and singularities, are more numerically stable, and are more computationally efficient than other representations \citep{pervin:quaternions:1983}. We provide a more thorough overview of rotation parameterizations in Section~\ref{sec:rotation_parameterization}.

The advantages of quaternions come at a cost: in order to represent valid rotations, they must be normalized to have unit length. To enforce this property, we add an explicit normalization layer to our network (cf. Figure~\ref{fig:architecture}). We also include a penalty term in the loss function, $\lambda (w^2 + x^2 + y^2 + z^2 - 1)^2$, for all quaternions prior to normalization. The latter acts as a regularizer and leads to better training stability. The choice of $\lambda$ is not crucial; we found that any value between $0.1$ and $0.001$ serves the purpose (we use $\lambda = 0.01$). During training, the distribution of the quaternion norms converges nicely to a Gaussian with mean 1, i.e. the model learns to represent valid rotations. It is important to observe that if $\mathbf{q}$ represents a particular orientation, then $\mathbf{-q}$ (\emph{antipodal representation}) represents the same orientation. 

\begin{figure*}
	\centering
	\subfigure[]{
		\includegraphics[height=0.225\textwidth]{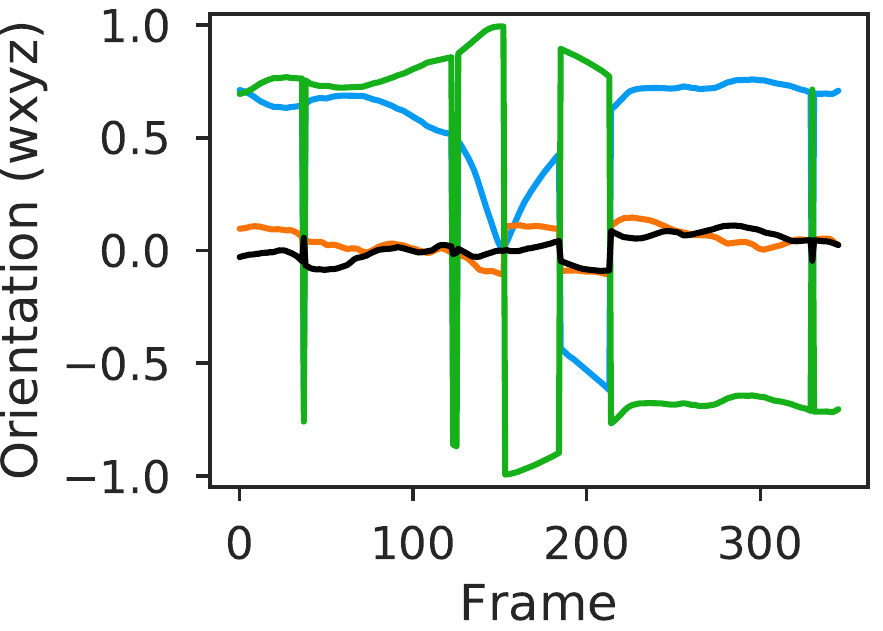}
		\label{fig:quaternions_antipodal}
	}\quad
	\subfigure[]{
		\includegraphics[height=0.225\textwidth]{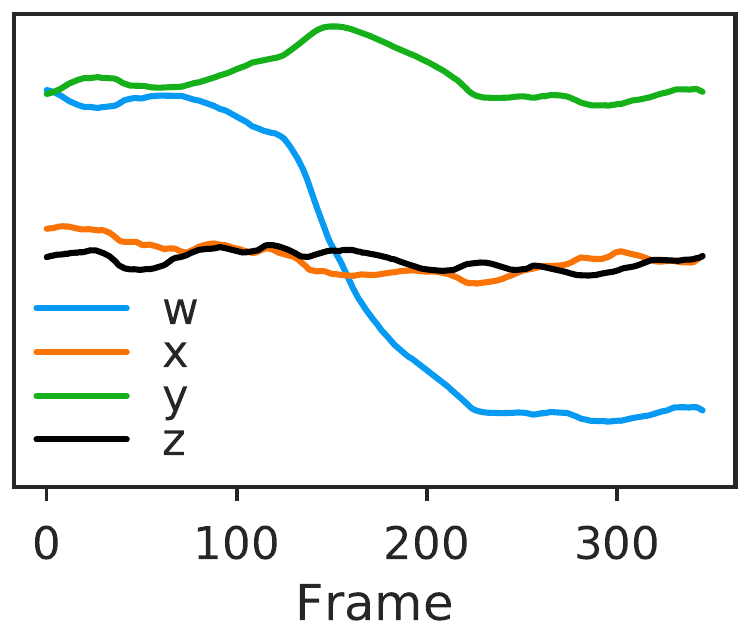}
		\label{fig:quaternions_antipodal_continuous}
	}\quad
	\subfigure[]{
     	\includegraphics[height=0.225\textwidth]{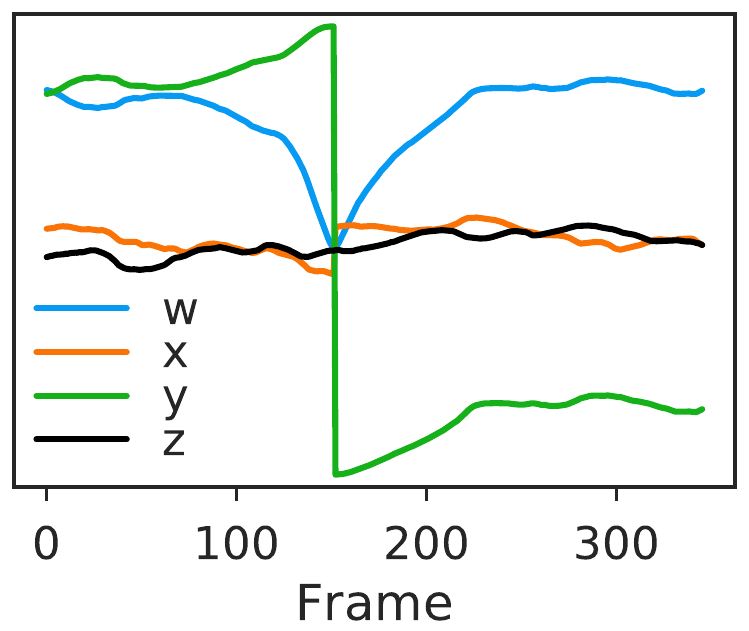}
 		\label{fig:quaternions_antipodal_unique}
    }
    \caption[]{
        Antipodal representation problem for quaternions.
        \textbf{\subref{fig:quaternions_antipodal}} A real sequence from the training set for the root joint rotation, both discontinuous and ambiguous.
        \textbf{\subref{fig:quaternions_antipodal_continuous}} Our approach, which corrects discontinuities but still allows for two possible choices, $\mathbf{q}$ and $-\mathbf{q}$.
        \textbf{\subref{fig:quaternions_antipodal_unique}} Unique representation obtained by forcing $w$ to be non-negative but which introduces discontinuities.
	}
	\label{fig:quaternions}
\end{figure*}

As shown in Figure~\ref{fig:quaternions_antipodal}, we found these two representations to be mixed in our dataset, leading to discontinuities in the time series. 
Our solution is to choose the representation with the lowest Euclidean distance (or equivalently, the highest cosine distance) from the one in the previous frame $t-1$ (Figure~\ref{fig:quaternions_antipodal_continuous}). 
This representation still allows for two representations with inverted sign for each time series, which does not represent an issue for autoregressive models.

Owing to the advantages presented above, this work represents joint rotations with quaternions.
Previous work in motion modeling has used quaternions for pose clustering \citep{zhou:quaternion:2013}, for joint limit estimation \citep{herda:quaternion_limit:2005}, and for motion retargeting \citep{villegas:cvpr:2018}. To the best of our knowledge, human motion prediction with a quaternion parameterization is a novel contribution of our work.

Discontinuities are not the only drawback of previous approaches (cf. Section~\ref{sec:related}). Regression of rotations fails to properly encode that a small error on a crucial joint might drastically impact the positional error. Therefore we propose to compute a positional loss. Our loss function takes as input joint rotations and runs forward kinematics to compute the position of each joint. We can then compute the Euclidean distance between each predicted joint position and the reference pose. Since forward kinematics are differentiable with respect to joint rotations, this is a valid loss for training the network. This approach is inspired by \cite{zhou:ijcai:2016} for hand tracking and \cite{zhou:eccvw:2016} for human pose estimation in static images. Unlike Euler angles (used in \cite{zhou:ijcai:2016,zhou:eccvw:2016}), which employ trigonometric functions to compute transformations, quaternion transformations are based on linear operators \citep{pervin:quaternions:1983} and are therefore more suited to neural network architectures. \cite{villegas:cvpr:2018} also employ a form of forward kinematics with quaternions, in which quaternions are converted to rotation matrices to compose transformations. In our case, all transformations are carried out in quaternion space and the network is conditioned on joint rotations, unlike \citep{villegas:cvpr:2018} which is conditioned on joint positions.
Compared to other work with positional loss \citep{holden:deeplearning:2016,holden:phase:2017}, our strategy penalizes position errors properly and avoids re-projection onto skeleton constraints. Additionally, our differentiable forward kinematics implementation allows for efficient GPU batching and therefore only increases the computational cost over the rotation-based loss by $\sim$20\%.

\subsection{Parameterization of rotations}
\label{sec:rotation_parameterization}

In this section, we compare different parameterizations for rotations in the 3D Euclidean space and we highlight their strengths and weaknesses in different contexts. 
All the presented representations model the 3D rotation group SO(3), which can be fully expressed with a minimum of 3 parameters.

\subsubsection{Euler angles}

They represent orientations as successive rotations around the axes of a coordinate system, typically referred to as \emph{yaw}, \emph{pitch}, and \emph{roll}. There are multiple ways to compose rotations and applications that use Euler angles must agree on the particular order convention: \emph{Tait-Bryan ordering} (xyz, xzy, yxz, yzx, zxy, zyx), or \emph{proper ordering} (xyx, xzx, yxy, yzy, zxz, zyz). 

A typical Euler rotation vector is a triplet that indicates the rotation around each axis in radians. 
There are two drawbacks: first, if $x$ represents a particular rotation, then $x + 2k\pi$ ($k \in \mathbb{Z}$) represents the same rotation. 
This means that there is an infinite number of representations for the same rotation. 
Moreover, the wrap-around issue at $2\pi$ causes the representation space to be discontinuous, which is undesirable in optimization or in applications that require smooth interpolation.

A trick to avoid the discontinuity issue with angles (whether 3D Euler angles or 1D angles) is to represent each angle $\theta$ as a 2D feature vector $[\cos\theta, \sin\theta]$, which is guaranteed to lie on the unit circle as $\cos^2\theta + \sin^2\theta = 1$. 
This can be equivalently viewed as a unit complex number $a + ib$. 
The corresponding approach to regress such angles would be to output two values $a$ and $b$, impose $a^2 + b^2 = 1$ either via a smooth constraint or via explicit normalization (or both, as we show in Section~\ref{sec:quaternion_loss} in the context of quaternions), and compute $\theta = \text{atan2}(b, a)$. 
This approach solves the discontinuity problem, but doubles the number of parameters, introduces an optimization constraint, and still presents no 3D interpolation properties.

As with other $\mathbb{R}^3$ parameterizations, Euler angles suffer from singularities. In the context of rotations, a singularity is a subspace in which all elements express the same rotation, which means that no rotation is possible within the subspace \citep{grassia:rotations:1998}. With Euler angles, this is referred to as \emph{gimbal lock}, and results in the loss of one degree of freedom due to the gimbals becoming ``interlocked'' -- an analogy with physical inertial measurement units (IMUs) based on Euler angles.

\subsubsection{Axis-angle representation}

Also referred to as the \emph{exponential map}, this representation again uses 3 parameters and is proposed as a more practical alternative to Euler angles. It mitigates some of the issues of the latter by making them unlikely \citep{grassia:rotations:1998}, but does not solve them at the fundamental level.

\begin{figure}
	\centering
	\includegraphics[width=0.25\textwidth]{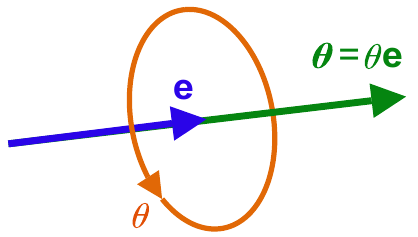}
	\caption{Rotation expressed in axis-angle representation.}
	\label{fig:axis-angle}
\end{figure}

Intuitively, an axis-angle rotation is described by an axis $\hat{\mathbf{e}}$ (a 3D vector $xyz$ with unit length which represents a direction), and a rotation angle $\theta$ around this axis. 
The latter is encoded as the length of the vector, i.e. $\theta = \sqrt{x^2 + y^2 + z^2}$.
This is shown in Figure~\ref{fig:axis-angle}. 
Singularities are present on every sphere of radius $2k\pi$, since they are equivalent to a rotation with $\theta = 0$. 
As with Euler angles, there are an infinite number of representations of the same rotation (one for each sphere). 
Even when restricting the parameter space to the sphere of radius $2\pi$, there are two possible representations: $(\hat{\mathbf{e}},\; \theta)$ and $(-\hat{\mathbf{e}},\; 2\pi - \theta)$. 
Likewise, the parameter space is discontinuous when $\theta$ wraps around from $2\pi$ to $0$.

Another disadvantage of exponential maps is that there is no way to compose rotations, even though it is possible to rotate vectors using \emph{Rodrigues' formula}~\citep{rodrigues}, which involves trigonometric functions. 
Composition is a fundamental operator for forward kinematics, and is trivial to achieve in rotation matrices (matrix multiplication) and quaternions (quaternion multiplication). \citet{grassia:rotations:1998} suggests to transform them to quaternions (the closest alternative), compose rotations, and convert them back to exponential maps, incurring several computations of trigonometric functions. 
\citet{grassia:rotations:1998} also observes that exponential maps are particularly suited to ball-and-socket joints, but they cannot be used for animating tumbling bodies. In human motion, one such an example is the root joint of a character spinning in circles, which has a range of motion greater than $2\pi$.

\subsubsection{Unit quaternions}

Quaternions are a 4D extension of complex numbers that form the $\mathbb{S}^3$ group, and can be described as real-valued 4-tuples $wxyz$ such that $\mathbf{q} = w + x\mathbf{i} + y\mathbf{j} + z\mathbf{k}$, where $w$ is the \emph{scalar term} and $xyz$ are the \emph{complex terms}. For rotations, we are interested in \emph{unit quaternions}, i.e. quaternions with unit length. A rotation of $\theta$ radians around an axis $\hat{\mathbf{v}}$ is encoded as $w = \cos(\theta/2)$ and $xyz = \hat{\mathbf{v}} \sin(\theta/2)$.

This representation is closely related to the exponential map -- describing a rotation around an axis -- but presents fundamental differences. It uses 4 parameters instead of 3, and requires the vector to be normalized (i.e. on the unit sphere). This small disadvantage compares to several advantages:
\begin{itemize}
    \item No singularities, since they are embedded in $\mathbb{R}^4$ and not $\mathbb{R}^3$.
    \item No discontinuities in the parameter space, which means that they can be regressed or interpolated smoothly.
    \item They can be composed and used to compute transformations without switching to other representations, and without requiring periodic functions.
    \item They present a simple and elegant way to perform interpolation between rotations (\emph{quaternion slerp}), which results in a continuous path and good qualitative properties such as constant velocity and minimal torque \citep{shoemake:quaternion:1985}. This respectively means that the artist has precise control over the transition speed, and that the transition is as smooth as possible.
\end{itemize}

A disadvantage of quaternions is that they encode half-angle rotations, giving rise to the so-called \emph{antipodal representations}: two possible representations for the same 3D orientation, $\mathbf{q}$ and $-\mathbf{q}$. Nevertheless, this dual representation is still advantageous compared to other parameterizations with infinite representations.

One approach to tackle this problem is to force $\mathbf{q}$ to cover only half of $\mathbb{S}^3$. For instance, a straightforward way of implementing this would be to require $w$ to be positive (i.e. inverting $\mathbf{q}$ if $w$ is negative). A more thorough approach would also consider the case of $w = 0$, and repeat the same process on $x$, and then on $y$ if necessary \citep{lavalle:planning:2006}. However, this trick causes the representation space to be discontinuous (see Figure~\ref{fig:quaternions_antipodal_unique} for an example), which defeats one of the main purposes of using quaternions.

In Section~\ref{sec:quaternion_loss}, we showed how we solved the antipodal representation problem in our data. 
Furthermore, the use of an autoregressive architecture allows the model to keep track of the current ``hemisphere'' in $\mathbb{S}^3$ and regress continuous rotations.

\subsection{Short-term prediction}
\label{sec:shortterm}

For short-term predictions with our quaternion network, we consider predicting either relative rotation deltas (analogous to angular velocities) or absolute rotations. We take inspiration from residual connections applied to Euler angles \citep{martinez:recurrent:2017}, where the model does not predict absolute angles but angle deltas and integrates them over time. For quaternions, the predicted deltas are applied to the input quaternions through quaternion product \citep{shoemake:quaternion:1985} (\emph{QMul} block in Figure~\ref{fig:architecture}). Similar to \cite{martinez:recurrent:2017}, we found this approach to be beneficial for short-term prediction, but we also discovered that it leads to instability for long-term generation.

\begin{figure}
	\centering
	\subfigure[\hspace{-8mm}]{
		\includegraphics[width=0.45\linewidth]{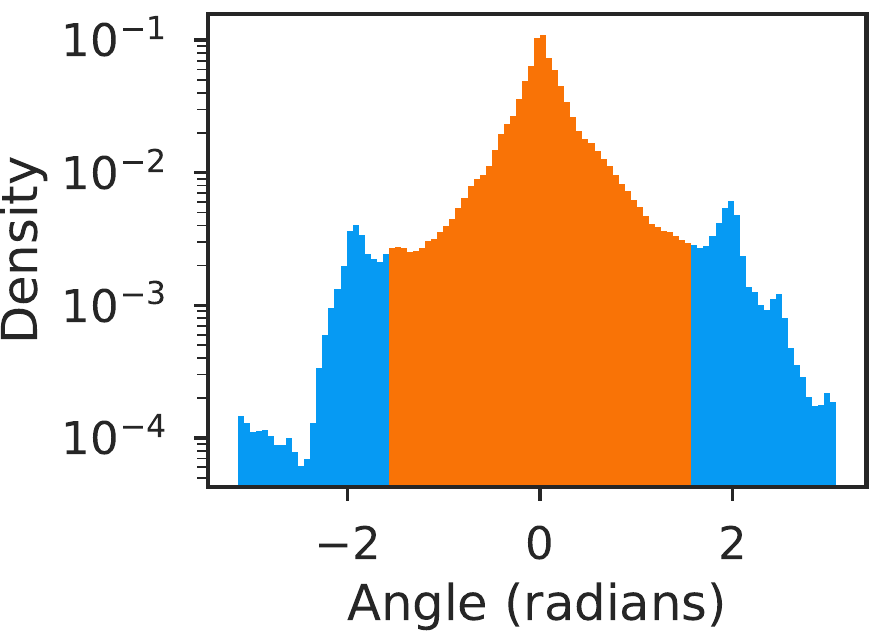}
		\label{fig:angle_hist}
	}\quad
	\subfigure[\hspace{-8mm}]{
		\includegraphics[width=0.45\linewidth]{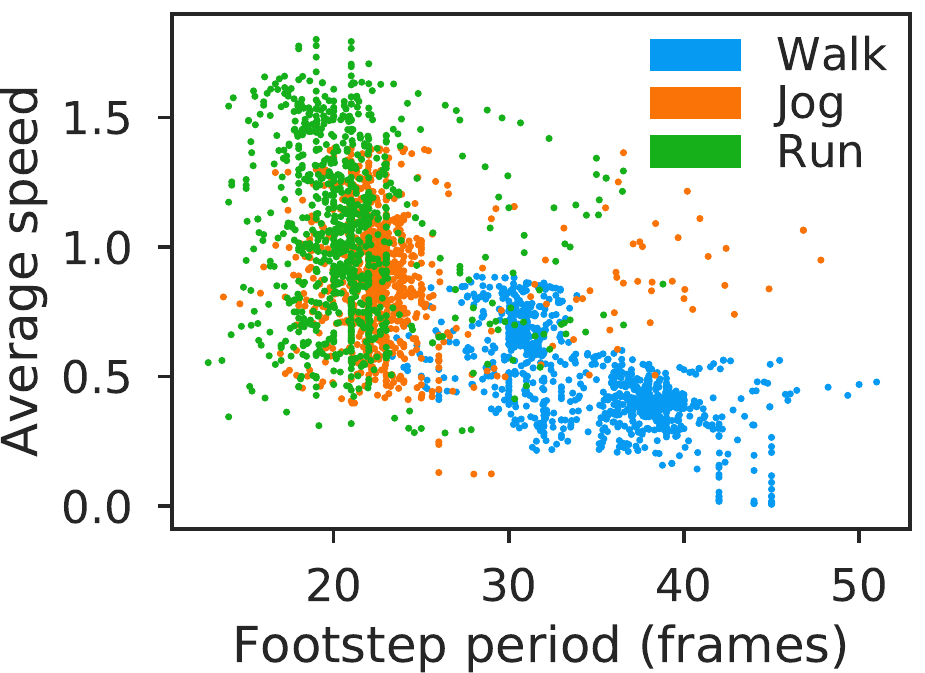}
		\label{fig:clusters}
	}
	\caption[]{
		\textbf{\subref{fig:angle_hist}} Local angle distribution for H3.6M, where orange represents the safe range between $-\pi/2$ and $\pi/2$, and blue highlights the potentially problematic range (7\% of all angles).
		\textbf{\subref{fig:clusters}} Distribution of the gait parameters across the training set of \cite{holden:deeplearning:2016}.
	}
	\label{fig:angles}
\end{figure}

Previous work evaluates prediction errors by measuring Euclidean distances between Euler angles and we precisely replicate that protocol to provide comparable results by replacing the positional loss with a loss on Euler angles.
This loss first maps quaternions onto Euler angles, and then computes the L1 distance with respect to the reference angles, taking the best match modulo $2\pi$. A proper treatment of angle periodicity was not found in previous implementations, e.g. \cite{martinez:recurrent:2017}, leading to slightly biased results. In particular, there is a non-neglible number of angles located around $\pm \pi$ in the dataset used for our experiments, see Figure~\ref{fig:angle_hist}.

\subsection{Long-term generation}
\label{sec:longterm}

For long-term generation, we restrict ourselves to locomotion actions. We define our task as the generation of a pose sequence given an average speed and a ground trajectory to follow. Such a task is common in computer graphics \citep{badler:simulating:1993,multon:animation:1999,forsyth:motion:2006}.

We decompose the task into two steps: we start by defining some parameters along the trajectory (facing direction of the character, local speed, frequency of footsteps), then we predict the sequence of poses. The trajectory parameters can be manually defined by the artist, or they can be fitted automatically via a simple \emph{pace network}, which is provided as a useful feature for generating an animation with minimal effort. The second step is addressed with our autoregressive quaternion network (\emph{pose network}).

The pace network is a simple recurrent network with one GRU layer with 30 hidden units. It represents the trajectory as a piecewise linear spline with equal-length segments \citep{Stoer:INA:1993} and performs its recursion over segments. At each time step, it receives the spline curvature and the previous hidden state. It predicts the character facing direction relative to the spline tangent (which can be used for making the character walk sideways, for instance), the frequency of its footsteps, and its local speed, which is a low-pass filtered version of the instantaneous speed on the training set. We found the two dimensions (frequency and speed) necessary to describe the character's gait (e.g. walk, jog, run), as illustrated in Figure~\ref{fig:clusters}.

This network is trained to minimize the mean absolute error (MAE) of its features. Depending on the scenario -- offline or online -- we propose two versions of this network: one based on a bidirectional architecture, and one based on a regular 1-directional RNN whose outputs are delayed by a small distance. The latter is particularly suitable for real-time applications, since it does not observe the trajectory far in the future.

The pose network is similar to the network we used for short-term predictions but presents additional inputs and outputs, i.e. the \emph{Translations} and \emph{Controls} blocks in Figure~\ref{fig:architecture}. The \emph{Controls} block consists of the tangent of the current spline segment as a 2D versor, the facing direction as a 2D versor, the local longitudinal speed along the spline, and the walk cycle. The last two features are merged into a signal of the form $A[\cos(\theta), \sin(\theta)]$, where $A$ is the longitudinal speed, and $\theta$ is a cyclic signal where $0 = 2\pi$ corresponds to a left foot contact and $\pi$ corresponds to a right foot contact. For training, we extract these features from training recordings by detecting when the speed of a foot falls to zero. At inference, we integrate the frequency to recover $\theta$. Since this block is not in the recurrent path, we pass its values through two fully connected layers with 30 units each and Leaky ReLU activations (with leakage factor $a = 0.05$). We use leaky activations to prevent the units from dying, which may represent a problem with such a small layer size. The pose network also takes the additional outputs from the previous time-step (\emph{Translations} block). These outputs are the height of the character root joint and the positional offset on the spline compared to the position obtained by integrating the average speed. The purpose of the latter is to model the high-frequency details of movement, which helps with realism and foot sliding. For training, we extract this feature from the training data by low-pass filtering the speed along the trajectory (which yields the average local speed), subtracting the latter from the overall speed (which yields a high-pass-filtered series), and integrating it. The pose network is trained to minimize the Euclidean distance to the reference pose with the forward kinematic positional loss introduced in Section~\ref{sec:quaternion_loss}. As before, we regularize non-normalized quaternion outputs to stay on the unit sphere.

\section{Experiments}
\label{sec:exp}

We perform two types of evaluation. We evaluate short-term prediction of human motion over different types of actions using the benchmark setting evaluating angle prediction errors on Human3.6M data \citep{fragkiadaki:recurent:2015,liu:spatiotemp:2016,martinez:recurrent:2017}. We also conduct a human study to qualitatively evaluate the long-term generation of human locomotion \citep{holden:deeplearning:2016,holden:phase:2017} since quantitative generation of long-term prediction is difficult. For the latter, we use the same dataset as \cite{holden:lmm:2015, holden:deeplearning:2016}, instead of Human3.6M. Finally, we perform various ablations in Section~ \ref{sec:ablations}, where we compare different rotation parameterizations and strategies.

\subsection{Short-term prediction}
\label{sec:shortterm_exp}

We follow the experimental setup of \cite{fragkiadaki:recurent:2015} on the Human3.6M task \citep{ionescu:iccv:2011, ionescu:human36:2014}. This dataset consists of motion capture data from seven actors performing 15 actions. The skeleton is represented with 32 joints recorded at 50 Hz, which we down-sample to 25 Hz keeping both even/odd versions of the data for training as in \cite{martinez:recurrent:2017}.
Our evaluation measures the Euclidean distance between predicted and measured Euler angles, similar to \citet{fragkiadaki:recurent:2015,liu:spatiotemp:2016,martinez:recurrent:2017}. We use the same train and test split, i.e. subjects 1, 6, 7, 8, 9, 11 for training, and subject 5 for testing. We compare to previous neural approaches \citep{fragkiadaki:recurent:2015,liu:spatiotemp:2016,martinez:recurrent:2017} and simple baselines \citep{martinez:recurrent:2017}: running average over 2 and 4 frames (Run. avg. 2/4) and zero-velocity which is the last known frame.

We train a single model for all actions, without supplying any action category as input. For the RNN architecture, we condition the generator on $n = 50$ frames (2 seconds) and predict the next $k = 10$ frames (400 ms). For the CNN architecture, we condition on $n = 32$ frames (1.28 s) and predict $k = 10$ frames (400 ms). We report results both for modeling velocities or relative rotations (\name vel.) and absolute rotations (\name abs.).
Table~\ref{tbl:h3.6m} shows the results and highlights that velocities generally perform better than absolute rotations for short-term predictions. 
It also shows that our RNN architecture performs better than the CNN architecture on this task and we therefore focus subsequent analysis on the RNN model.

To better understand the effect of scheduled sampling, we also train a model without scheduled sampling and without feedback, i.e., teacher forcing (\name vel. TF). In this setting we compute the loss directly on quaternions instead of Euler angles, to enforce their continuity. We define the similarity of two quaternions $\mathbf{p}$ and $\mathbf{q}$ as their dot product, resulting in the loss function:
$$
    E(\mathbf{p},\, \mathbf{q}) = 1 - \mathbf{p} \cdot \mathbf{q} .
$$
This error also corresponds to half the Euclidean distance, i.e. root mean square error, since quaternions have unit norm.
On the recurrent model, this experiment shows that teacher forcing achieves a slightly lower error on shorter time spans (80 ms) but does worse than scheduled sampling 
for longer time spans.
Exposing the model to the actual predictions at training time makes it less susceptible to diverging over longer time horizons. 
Interestingly, scheduled sampling seems much less effective for the convolutional model.

\begin{figure}
\centering
    \subfigure[\hspace{-5mm}]{
    	\includegraphics[width=0.46\linewidth]{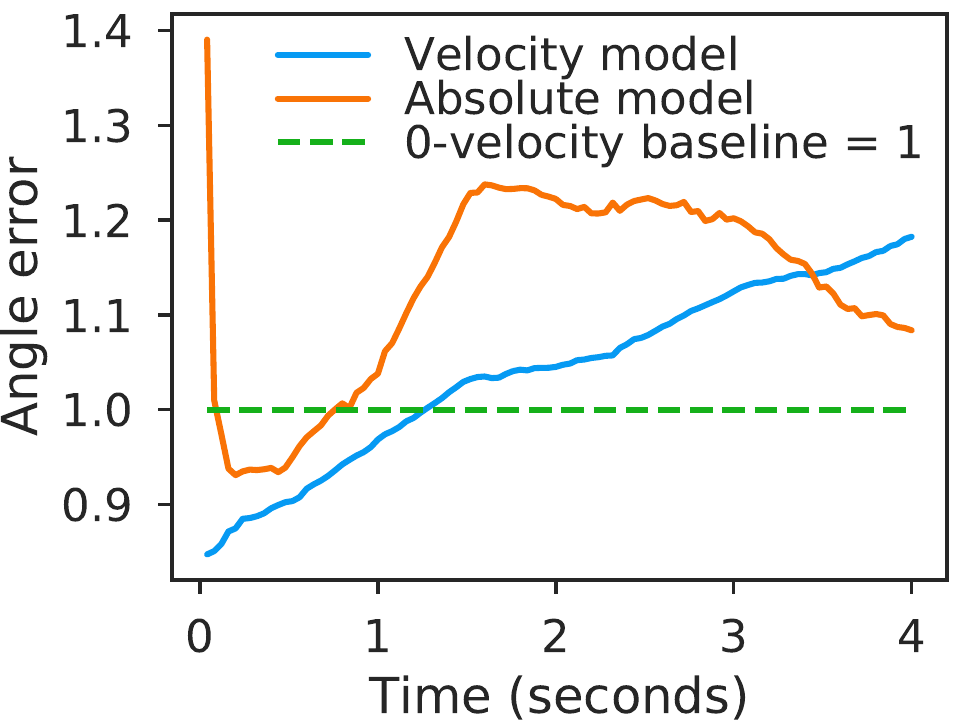}
		\label{fig:velocity_vs_orientation}
    }
    \subfigure[\hspace{-5mm}]{
    	\includegraphics[width=0.46\linewidth]{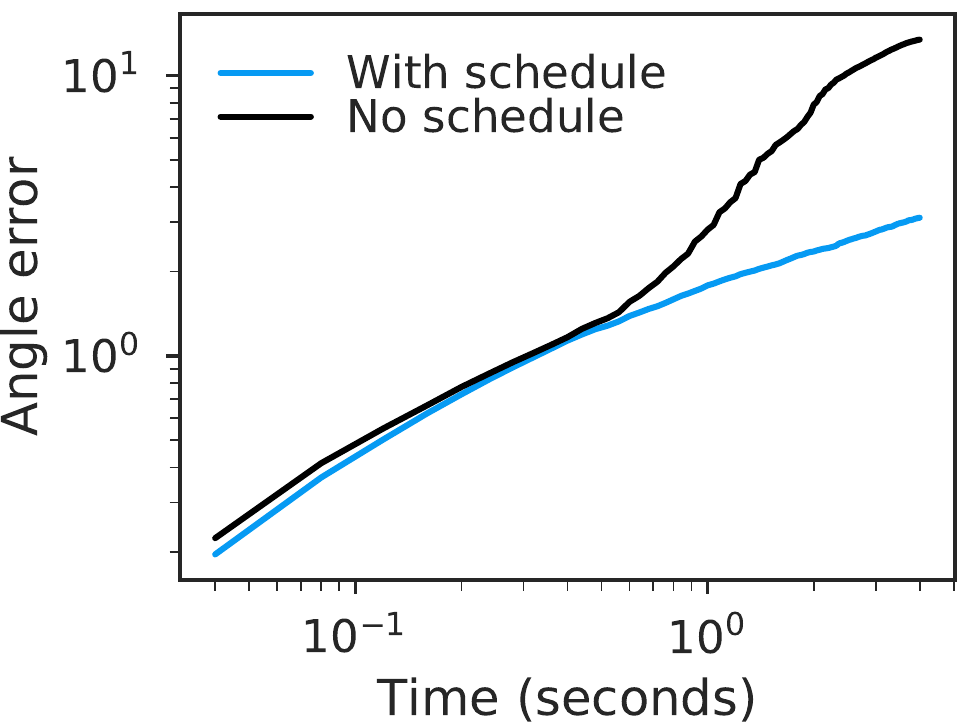}
		\label{fig:schedule_vs_sampling}
    }
\caption{Comparison between models for a longer time span, using the recurrent architecture. We compare the mean angle errors for all 15 actions, each averaged over 64 test sequences.
\textbf{\subref{fig:velocity_vs_orientation}} Velocity model vs orientation model, with respect to the zero-velocity baseline (for clarity). Both models are trained with scheduled sampling.
\textbf{\subref{fig:schedule_vs_sampling}} Beneficial effect of training with scheduled sampling on the velocity model.
}
\label{fig:longer}
\end{figure}

We report results with a longer-term horizon on all 15 actions. Figure~\ref{fig:velocity_vs_orientation} shows that integrating velocities is prone to error accumulation and absolute rotations are therefore advantageous for longer-term predictions. The graph also highlights that motion becomes mostly stochastic after the 1-second mark, and that the absolute rotation model presents small discontinuities when the first frame is predicted, which corroborates the findings of \cite{martinez:recurrent:2017}. Figure~\ref{fig:schedule_vs_sampling} reveals that if the recurrent velocity model is trained with scheduled sampling, it tends to learn a more stable behavior for long-term predictions. By contrast, the velocity model trained with regular feedback is prone to catastrophic drifts over time.

\begin{table*}[t!]
\centering
\small
\resizebox{\textwidth}{!}{
\tabcolsep=1mm
{\scalefont{1.2}
\begin{tabular}{@{}lrrrr|rrrr|rrrr|rrrr@{}}
 & \multicolumn{4}{c}{Walking} & \multicolumn{4}{c}{Eating} & \multicolumn{4}{c}{Smoking} & \multicolumn{4}{c}{Discussion}\\
milliseconds & 80 & 160 & 320 & 400 & 80 & 160 & 320 & 400 & 80 & 160 & 320 & 400 & 80 & 160 & 320 & 400 \\
\midrule
Run. avg. 4 \citeconf{martinez:recurrent:2017}{CVPR} & 0.64 & 0.87 & 1.07 & 1.20 & 0.40 & 0.59 & 0.77 & 0.88 & 0.37 & 0.58 & 1.03 & 1.02 & 0.60 & 0.90 & 1.11 & 1.15\\
Run. avg. 2 \citeconf{martinez:recurrent:2017}{CVPR} & 0.48 & 0.74 & 1.02 & 1.17 & 0.32 & 0.52 & 0.74 & 0.87 & 0.30 & 0.52 & 0.99 & 0.97 & 0.41 & 0.74 & 0.99 & 1.09\\
Zero-velocity \citeconf{martinez:recurrent:2017}{CVPR} & 0.39 & 0.68 & 0.99 & 1.15 & 0.27 & 0.48 & 0.73 & 0.86 & 0.26 & 0.48 & 0.97 & 0.95 & 0.31 & 0.67 & 0.94 & 1.04\\
\midrule
ERD \citeconf{fragkiadaki:recurent:2015}{CVPR}       & 0.93 & 1.18 & 1.59 & 1.78 & 1.27 & 1.45 & 1.66 & 1.80 & 1.66 & 1.95 & 2.35 & 2.42 & 2.27 & 2.47 & 2.68 & 2.76\\
LSTM-3LR \citeconf{fragkiadaki:recurent:2015}{CVPR}  & 0.77 & 1.00 & 1.29 & 1.47 & 0.89 & 1.09 & 1.35 & 1.46 & 1.34 & 1.65 & 2.04 & 2.16 & 1.88 & 2.12 & 2.25 & 2.23\\
SRNN \citeconf{jain:srnn:2016}{CVPR}             & 0.81 & 0.94 & 1.16 & 1.30 & 0.97 & 1.14 & 1.35 & 1.46 & 1.45 & 1.68 & 1.94 & 2.08 & 1.22 & 1.49 & 1.83 & 1.93\\
GRU unsup. \citeconf{martinez:recurrent:2017}{CVPR}         &  0.27 &  0.47 &  0.70 &  0.78 & 0.25 & 0.43 & 0.71 & 0.87 & 0.33 & 0.61 & 1.04 & 1.19 &  0.31 & 0.69 & 1.03 & 1.12\\
GRU sup. \citeconf{martinez:recurrent:2017}{CVPR}           & 0.28 & 0.49 & 0.72 & 0.81 &  0.23 &  0.39 &  0.62 &  0.76 & 0.33 & 0.61 & 1.05 & 1.15 &  0.31 & 0.68 & 1.01 & 1.09\\
Adversarial \citeconf{gui:adversarial:2018}{ECCV} & 0.22 & \underline{0.36} & {\bf 0.55} & 0.67 & {\bf 0.17} & {\bf 0.28} & {\bf 0.51} & {\bf 0.64} & 0.27 & {\bf 0.43} & {\bf 0.82} & {\bf 0.84} & 0.27 & {\bf 0.56} & {\bf 0.76} & {\bf 0.83}\\
\midrule

\name abs. \citeconf{pavllo:quaternet_bmvc:2018}{BMVC} & 0.26 & 0.42 & 0.67 & 0.70    &    0.23 & 0.38 & 0.61 & 0.73    &    0.32 & 0.52 & 0.92 & \underline{0.90}    & 0.36 & 0.71 & 0.96 & 1.03 \\
\name vel. \citeconf{pavllo:quaternet_bmvc:2018}{BMVC}  &  \underline{0.21} & {\bf 0.34} & \underline{0.56} & {\bf 0.62}    &    0.20 & 0.35 & \underline{0.58} & \underline{0.70}    &    \underline{0.25} & \underline{0.47} & 0.93 & \underline{0.90}    & \underline{0.26} & \underline{0.60} & \underline{0.85} & \underline{0.93} \\
\name vel. TF & {\bf 0.20} & 0.37 & 0.64 & 0.76 & \underline{0.19} & \underline{0.34} & 0.61 & 0.78 & {\bf 0.24} & 0.48 & \underline{0.90} & 0.99 & {\bf 0.25} & 0.64 & 0.97 & 1.07 \\
\midrule
\name CNN abs. & 0.31 & 0.61 & 0.89 & 0.96 & 0.27 & 0.54 & 0.86 & 1.02 & 0.37 & 0.76 & 1.26 & 1.33 & 0.38 & 0.84 & 1.16 & 1.22   \\
\name CNN vel. & 0.25 & 0.40 & 0.62 & 0.70 & 0.22 & 0.36 & \underline{0.58} & 0.71 & 0.26 & 0.49 & 0.94 & \underline{0.90} & 0.30 & 0.66 & 0.93 & 1.00 \\
\name CNN vel. TF & \underline{0.21} & 0.39 & 0.65 & 0.75 & 0.20 & 0.36 & 0.65 & 0.83 & 0.26 & 0.49 & 0.96 & 1.07 & 0.30 & 0.67 & 0.99 & 1.09  \\

\bottomrule
\end{tabular}
}}
\caption{Results under the \textbf{standard protocol} \citep{fragkiadaki:recurent:2015}, with 4 samples per sequence. We shows the mean angle error for short-term motion prediction on Human 3.6M for different actions: simple baselines (top), previous RNN results (middle), \name (bottom). Bold indicates the best result, underlined indicates the second best. abs. = model absolute rotations, vel. = model velocities, TF = teacher forcing.}
\label{tbl:h3.6m}
\end{table*}

\begin{table*}[t!]
\centering
\small
\resizebox{\textwidth}{!}{
\tabcolsep=0.7mm
{\scalefont{1.2}
\begin{tabular}{@{}lrrrr|rrrr|rrrr|rrrr|rrrr|rrrr|rrrr|rrrr@{}}
 & \multicolumn{4}{c}{Walking} & \multicolumn{4}{c}{Eating} & \multicolumn{4}{c}{Smoking} & \multicolumn{4}{c}{Discussion} & \multicolumn{4}{c}{Directions} & \multicolumn{4}{c}{Greeting} & \multicolumn{4}{c}{Phoning} & \multicolumn{4}{c}{Posing}\\
milliseconds & 80 & 160 & 320 & 400 & 80 & 160 & 320 & 400 & 80 & 160 & 320 & 400 & 80 & 160 & 320 & 400 & 80 & 160 & 320 & 400 & 80 & 160 & 320 & 400 & 80 & 160 & 320 & 400 & 80 & 160 & 320 & 400 \\
\midrule
Run. avg. 4 & 0.64 & 0.92 & 1.30 & 1.39 & 0.46 & 0.69 & 0.98 & 1.09 & 0.48 & 0.67 & 1.02 & 1.14 & 0.74 & 1.00 & 1.35 & 1.46 & 0.46 & 0.67 & 0.99 & 1.14 & 0.94 & 1.20 & 1.56 & 1.69 & 0.60 & 0.84 & 1.23 & 1.37 & 0.64 & 0.93 & 1.35 & 1.54 \\
Run. avg. 2 & 0.51 & 0.83 & 1.26 & 1.36 & 0.35 & 0.63 & 0.95 & 1.07 & 0.37 & 0.59 & 0.96 & 1.08 & 0.60 & 0.90 & 1.31 & 1.45 & 0.36 & 0.59 & 0.95 & 1.10 & 0.78 & 1.10 & 1.51 & 1.66 & 0.48 & 0.75 & 1.18 & 1.33 & 0.50 & 0.82 & 1.29 & 1.48 \\
Zero-velocity & 0.43 & 0.78 & 1.23 & 1.34 & 0.30 & 0.59 & 0.94 & 1.07 & 0.34 & 0.56 & 0.94 & 1.08 & 0.55 & 0.83 & 1.27 & 1.46 & 0.30 & 0.54 & 0.92 & 1.08 & 0.67 & 1.03 & 1.47 & 1.66 & 0.42 & 0.71 & 1.17 & 1.31 & 0.42 & 0.75 & 1.25 & 1.45 \\
\midrule
GRU unsup. & 0.34 & 0.61 & 0.92 & 1.02 & 0.32 & 0.60 & 0.92 & 1.05 & 0.43 & 0.79 & 1.15 & 1.31 & 0.57 & 0.88 & 1.34 & 1.48 & 0.32 & 0.58 & 0.98 & 1.15 & 0.66 & 0.98 & 1.41 & 1.55 & 0.43 & 0.71 & 1.14 & 1.31 & 0.47 & 0.84 & 1.39 & 1.58  \\
GRU sup. & 0.34 & 0.60 & 0.91 & 0.98 & 0.30 & 0.57 & 0.87 & 0.98 & 0.35 & 0.69 & 1.14 & 1.29 & 0.54 & 0.85 & 1.30 & 1.44 & 0.32 & 0.58 & 0.97 & 1.14 & 0.64 & 0.99 & 1.40 & 1.54 & 0.42 & 0.70 & 1.11 & 1.27 & 0.46 & 0.83 & 1.33 & 1.52  \\

\midrule

\name abs. & 0.35 & 0.56 & 0.84 & 0.92 & 0.29 & 0.52 & 0.79 & 0.89 & 0.52 & 0.68 & 0.95 & 1.06 & 0.54 & 0.86 & 1.24 & 1.44 & 0.27 & 0.47 & 0.84 & 1.00 & 0.54 & 0.85 & 1.27 & 1.47 & 0.40 & 0.62 & 0.99 & 1.14 & 0.48 & 0.75 & 1.17 & 1.36 \\
\name vel. & 0.28 & 0.49 & 0.76 & 0.83 & 0.22 & 0.47 & 0.76 & 0.88 & 0.28 & 0.47 & 0.79 & 0.91 & 0.48 & 0.74 & 1.20 & 1.37 & 0.24 & 0.46 & 0.84 & 1.01 & 0.61 & 0.93 & 1.34 & 1.51 & 0.36 & 0.61 & 0.98 & 1.14 & 0.38 & 0.71 & 1.20 & 1.39  \\
\name vel. TF & 0.27 & 0.51 & 0.83 & 0.93 & 0.22 & 0.50 & 0.86 & 0.99 & 0.28 & 0.53 & 0.97 & 1.15 & 0.49 & 0.79 & 1.25 & 1.41 & 0.23 & 0.48 & 0.92 & 1.10 & 0.55 & 0.87 & 1.32 & 1.51 & 0.36 & 0.62 & 1.04 & 1.21 & 0.34 & 0.69 & 1.21 & 1.44 \\
\midrule
\name CNN abs. & 0.39 & 0.77 & 1.12 & 1.21 & 0.34 & 0.73 & 1.11 & 1.22 & 0.63 & 0.97 & 1.28 & 1.43 & 0.64 & 1.06 & 1.54 & 1.70 & 0.36 & 0.72 & 1.16 & 1.34 & 0.72 & 1.11 & 1.54 & 1.68 & 0.48 & 0.85 & 1.32 & 1.49 & 0.60 & 1.06 & 1.59 & 1.79 \\
\name CNN vel. & 0.31 & 0.54 & 0.83 & 0.91 & 0.27 & 0.53 & 0.81 & 0.92 & 0.31 & 0.51 & 0.92 & 1.04 & 0.52 & 0.83 & 1.24 & 1.42 & 0.29 & 0.53 & 0.90 & 1.06 & 0.66 & 1.00 & 1.41 & 1.58 & 0.39 & 0.63 & 1.03 & 1.19 & 0.41 & 0.74 & 1.24 & 1.44
 \\
\name CNN vel. TF & 0.29 & 0.53 & 0.87 & 0.97 & 0.23 & 0.51 & 0.87 & 1.01 & 0.29 & 0.51 & 0.90 & 1.07 & 0.49 & 0.82 & 1.35 & 1.65 & 0.24 & 0.50 & 0.93 & 1.12 & 0.57 & 0.90 & 1.36 & 1.56 & 0.37 & 0.64 & 1.10 & 1.30 & 0.37 & 0.72 & 1.25 & 1.48 \\
\midrule\midrule
\\
 & \multicolumn{4}{c}{Purchases} & \multicolumn{4}{c}{Sitting} & \multicolumn{4}{c}{Sitting Down} & \multicolumn{4}{c}{Taking Photo} & \multicolumn{4}{c}{Waiting} & \multicolumn{4}{c}{Walk Dog} & \multicolumn{4}{c}{Walk Together} & \multicolumn{4}{c}{Average}\\
milliseconds & 80 & 160 & 320 & 400 & 80 & 160 & 320 & 400 & 80 & 160 & 320 & 400 & 80 & 160 & 320 & 400 & 80 & 160 & 320 & 400 & 80 & 160 & 320 & 400 & 80 & 160 & 320 & 400 & 80 & 160 & 320 & 400 \\
\midrule
Run. avg. 4 & 0.80 & 1.09 & 1.41 & 1.50 & 0.57 & 0.81 & 1.15 & 1.28 & 0.72 & 1.01 & 1.45 & 1.62 & 0.39 & 0.54 & 0.80 & 0.90 & 0.57 & 0.82 & 1.24 & 1.39 & 0.74 & 0.97 & 1.27 & 1.34 & 0.57 & 0.79 & 1.08 & 1.18 & 0.62 & 0.86 & 1.21 & 1.34 \\
Run. avg. 2 & 0.66 & 1.01 & 1.38 & 1.47 & 0.45 & 0.71 & 1.09 & 1.22 & 0.59 & 0.90 & 1.37 & 1.54 & 0.31 & 0.48 & 0.75 & 0.87 & 0.45 & 0.71 & 1.17 & 1.32 & 0.61 & 0.90 & 1.23 & 1.32 & 0.46 & 0.72 & 1.05 & 1.17 & 0.50 & 0.78 & 1.16 & 1.30 \\
Zero-velocity & 0.57 & 0.96 & 1.36 & 1.45 & 0.38 & 0.65 & 1.04 & 1.18 & 0.51 & 0.85 & 1.33 & 1.51 & 0.26 & 0.44 & 0.73 & 0.84 & 0.39 & 0.64 & 1.13 & 1.28 & 0.53 & 0.85 & 1.20 & 1.31 & 0.40 & 0.67 & 1.03 & 1.15 & 0.43 & 0.72 & 1.13 & 1.28 \\
\midrule
GRU unsup. & 0.57 & 0.97 & 1.38 & 1.48 & 0.42 & 0.77 & 1.24 & 1.43 & 0.60 & 1.03 & 1.68 & 1.92 & 0.31 & 0.53 & 0.90 & 1.06 & 0.42 & 0.70 & 1.25 & 1.44 & 0.52 & 0.85 & 1.22 & 1.33 & 0.37 & 0.60 & 0.89 & 1.00 & 0.45 & 0.76 & 1.19 & 1.34 \\
GRU sup. & 0.57 & 0.95 & 1.33 & 1.43 & 0.41 & 0.75 & 1.22 & 1.41 & 0.59 & 1.00 & 1.62 & 1.87 & 0.30 & 0.52 & 0.88 & 1.02 & 0.41 & 0.68 & 1.20 & 1.37 & 0.52 & 0.84 & 1.21 & 1.32 & 0.35 & 0.57 & 0.83 & 0.94 & 0.43 & 0.74 & 1.15 & 1.30 \\

\midrule

\name abs. & 0.51 & 0.86 & 1.31 & 1.42 & 0.47 & 0.66 & 1.07 & 1.20 & 0.76 & 0.98 & 1.38 & 1.57 & 0.34 & 0.47 & 0.74 & 0.86 & 0.43 & 0.65 & 1.04 & 1.19 & 0.50 & 0.77 & 1.12 & 1.23 & 0.31 & 0.49 & 0.75 & 0.85 & 0.45 & 0.68 & 1.03 & 1.17  \\
\name vel. & 0.54 & 0.92 & 1.36 & 1.47 & 0.34 & 0.59 & 1.00 & 1.15 & 0.47 & 0.81 & 1.31 & 1.50 & 0.23 & 0.39 & 0.69 & 0.81 & 0.32 & 0.54 & 1.00 & 1.15 & 0.48 & 0.78 & 1.12 & 1.21 & 0.28 & 0.45 & 0.69 & 0.79 & 0.37 & 0.62 & 1.00 & 1.14 \\
\name vel. TF & 0.47 & 0.87 & 1.33 & 1.44 & 0.32 & 0.60 & 1.03 & 1.19 & 0.48 & 0.85 & 1.45 & 1.70 & 0.23 & 0.42 & 0.78 & 0.93 & 0.32 & 0.58 & 1.11 & 1.30 & 0.45 & 0.77 & 1.13 & 1.23 & 0.27 & 0.48 & 0.78 & 0.91 & 0.35 & 0.64 & 1.07 & 1.23 \\
\midrule
\name CNN abs. & 0.62 & 1.09 & 1.54 & 1.66 & 0.58 & 1.05 & 1.64 & 1.84 & 0.92 & 1.52 & 2.08 & 2.29 & 0.38 & 0.73 & 1.13 & 1.29 & 0.53 & 0.96 & 1.50 & 1.67 & 0.57 & 0.97 & 1.38 & 1.51 & 0.38 & 0.67 & 1.01 & 1.15 & 0.54 & 0.95 & 1.40 & 1.55 \\
\name CNN vel. & 0.56 & 0.94 & 1.34 & 1.43 & 0.35 & 0.63 & 1.04 & 1.18 & 0.51 & 0.85 & 1.33 & 1.51 & 0.26 & 0.44 & 0.74 & 0.86 & 0.37 & 0.61 & 1.07 & 1.22 & 0.50 & 0.80 & 1.14 & 1.24 & 0.31 & 0.51 & 0.76 & 0.88 & 0.40 & 0.67 & 1.05 & 1.19 \\
\name CNN vel. TF & 0.49 & 0.90 & 1.38 & 1.50 & 0.34 & 0.63 & 1.12 & 1.33 & 0.51 & 0.91 & 1.56 & 1.88 & 0.25 & 0.45 & 0.82 & 0.99 & 0.33 & 0.59 & 1.12 & 1.32 & 0.48 & 0.80 & 1.17 & 1.29 & 0.29 & 0.51 & 0.83 & 0.96 & 0.37 & 0.66 & 1.11 & 1.30 \\

\bottomrule
\end{tabular}
}
}
\caption{Results under \textbf{our proposed protocol}, with 128 samples per sequence compared to 4 samples as in Table~\ref{tbl:h3.6m}. We show the error for all 15 actions, as well as the average across actions.}
\label{tbl:h3.6m_protocol2}
\end{table*}

\subsection{More consistent short-term evaluation}

The standard evaluation protocol of~\citet{fragkiadaki:recurent:2015} constructs the test set by sampling random chunks from the test animations. 
This has the advantage of requiring much less computation than evaluating the loss over all possible subsequences.
The reference implementation samples only four chunks from each test sequence at random positions, using a fixed seed to initialize the random generator\footnote{Reference implementation at \url{https://github.com/asheshjain399/RNNexp/blob/srnn/structural_rnn/forecastTrajectories.py#L29}}. 
This exact methodology is adopted by \cite{liu:spatiotemp:2016, martinez:recurrent:2017, pavllo:quaternet_bmvc:2018, gui:adversarial:2018} and makes the quantitative results across these papers comparable.

However, using only four samples results in a very high variance of the test results as we show next.
This is especially evident when comparing results from different initialization seeds. It is also a concern for comparisons with the same seed, since the samples are not large enough to be representative of the whole test set. 
It causes slightly biased results, and most importantly, it makes it hard to reliably compare different architectures. 

\begin{figure}
    \centering
    \subfigure[``Walking'' after 80 ms]{
    	\includegraphics[width=0.46\linewidth]{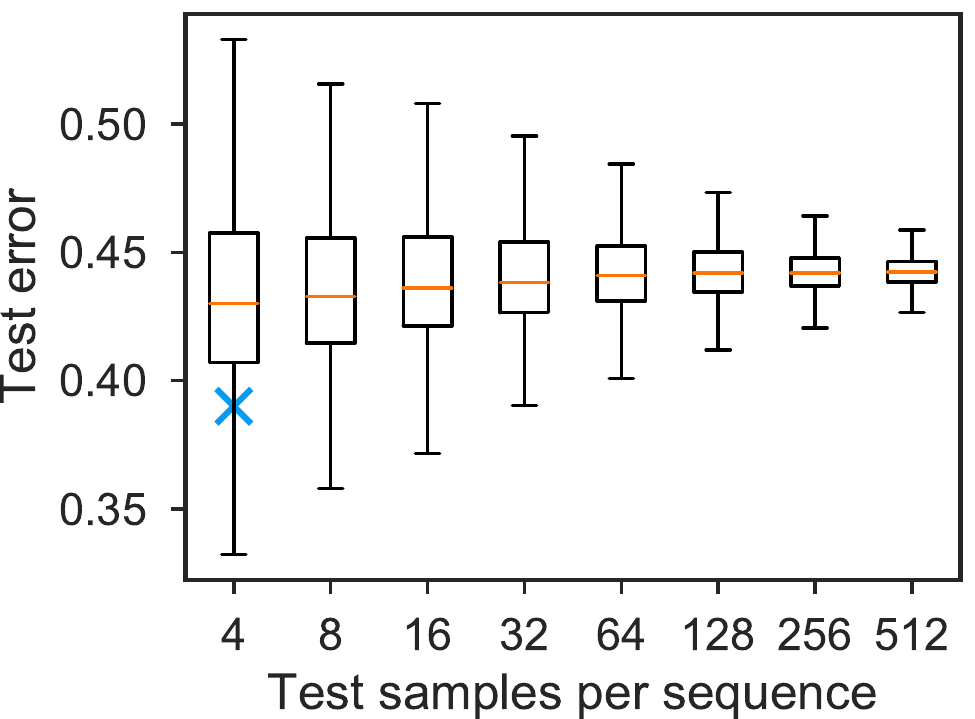}
		\label{fig:uncertainty_walking}
    }
    \subfigure[Average after 80 ms]{
    	\includegraphics[width=0.46\linewidth]{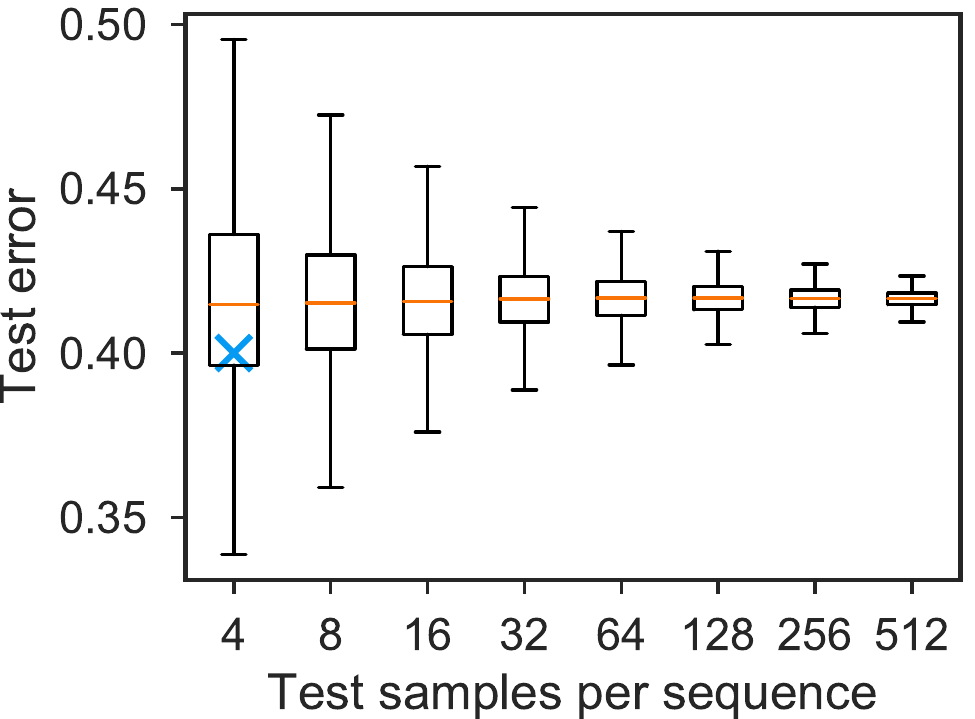}
		\label{fig:uncertainty_average}
    }
    \caption{
    Effect of increasing the number of samples per test sequence from the standard protocol of 4 to 512.
    We compute confidence intervals over the test error by bootstrap resampling of a large number of runs with different seeds.
    Results are based on the zero-velocity baseline for ``Walking'' and averaging over all 15 actions. 
    Small crosses denote the error corresponding to the default seed by \cite{fragkiadaki:recurent:2015}.
    }
    \label{fig:uncertainty}
\end{figure}

To quantify the issue, we compute the zero-velocity baseline \citep{martinez:recurrent:2017} for an increasing number of samples per sequence. 
Figure~\ref{fig:uncertainty} shows that four samples per sequence are not enough, since the error can vary by 10\% (0.395 -- 0.435) between the 25th and 75th quantile for the average over all actions (Figure~\ref{fig:uncertainty_average}).
This range can be reduced to 1.7\% (0.413 -- 0.420) with 128 samples, a number we believe to be a good compromise between variance and computational effort.

Finally, we compare different approaches under the new protocol. 
We also re-evaluated the approach of \cite{martinez:recurrent:2017} (GRU unsup./sup.) on all 15 actions by changing only the number of samples in their public implementation, we kept the same seed. 
The results for the new protocol (Table~\ref{tbl:h3.6m_protocol2}) show that the standard protocol tends to underestimate the true error (cf. Table~\ref{tbl:h3.6m}).
Moreover, it becomes easier to compare different strategies as any differences are less effected by noise.

\subsection{Long-term generation}
\label{sec:longterm_exp}

Our long-term evaluation relies on the generation of locomotion sequences from a given trajectory. We follow the setting of \cite{holden:deeplearning:2016}. The training set comprises motion capture data from multiple sources \citep{cmu:mocap,muller:hdm05:2007,ofli:mhad:2013,xia:style:2015} at 120 Hz, and is re-targeted to a common skeleton. In our case, we trained at a frame rate of 30Hz, keeping all 4 down-sampled versions of the data, and mirroring the skeleton to double the amount of data. We also applied random rotations to the whole trajectory to better cover the space of the root joint orientations. This dataset relies on the CMU skeleton \citep{cmu:mocap} with 31 joints. We removed joints with constant angle, yielding a dataset with 26 joints.

\begin{figure*}[t]
    \centering
    \subfigure[\hspace{-6mm}]{
    	\includegraphics[height=0.17\textwidth]{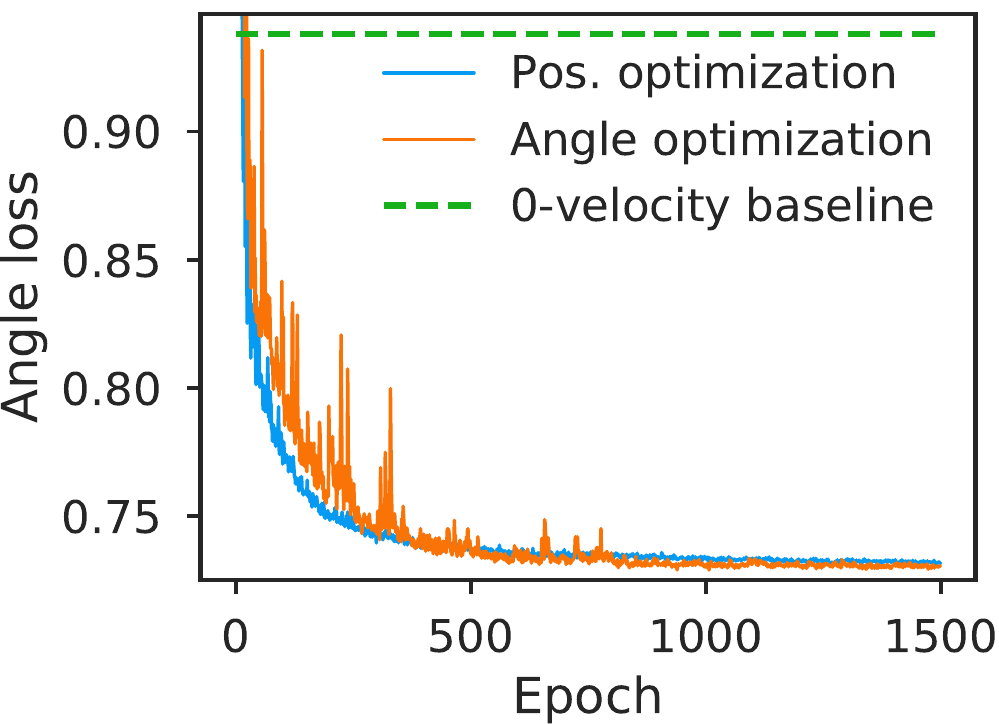}
		\label{fig:angle_loss}
    }
    \subfigure[\hspace{-5mm}]{
    	\includegraphics[height=0.17\textwidth]{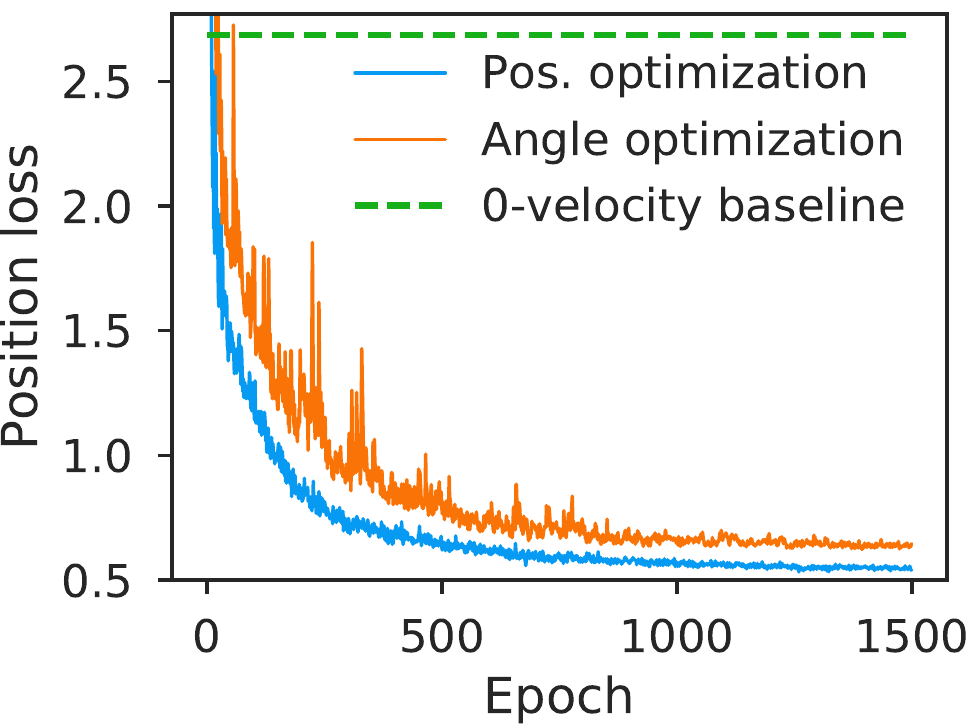}
		\label{fig:position_loss}
    }
    \subfigure[\hspace{-5mm}]{
    	\includegraphics[height=0.17\textwidth]{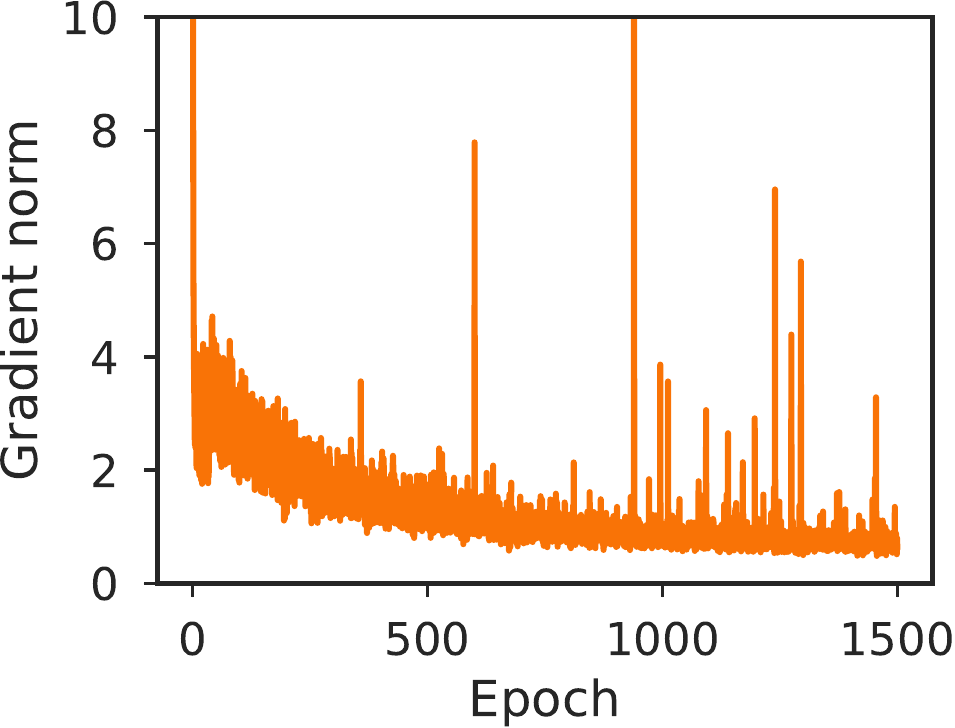}
		\label{fig:angle_loss_gradient}
    }
    \subfigure[\hspace{-5mm}]{
    	\includegraphics[height=0.17\textwidth]{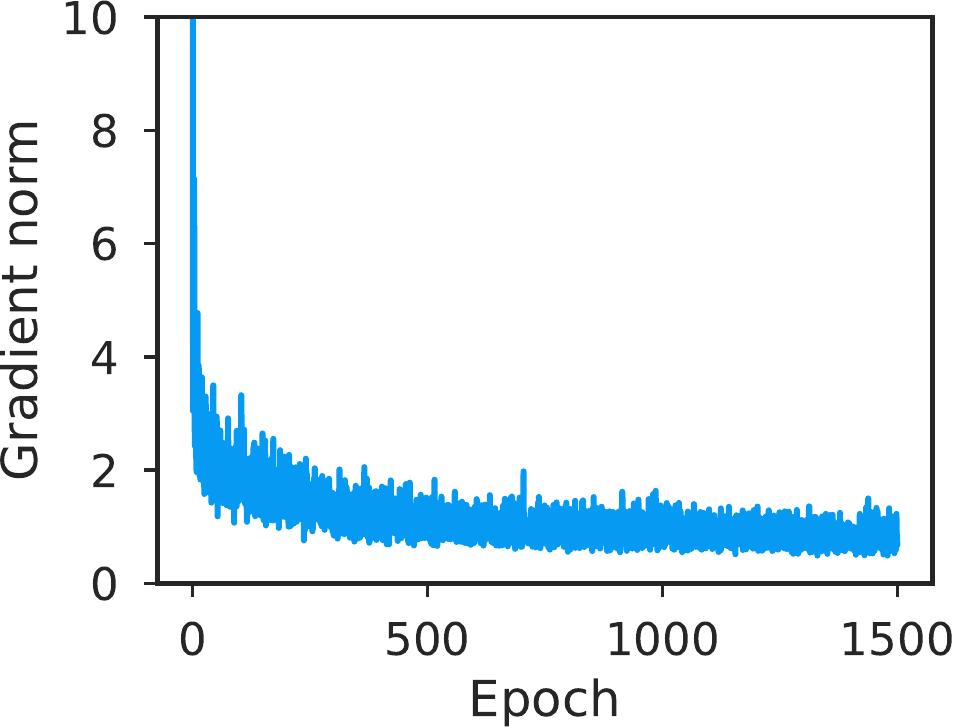}
		\label{fig:position_loss_gradient}
    }
    \caption{Training with angle versus positional loss on long-term generation.
      \textbf{\subref{fig:angle_loss}} Angle distance between joint orientations.
      \textbf{\subref{fig:position_loss}} Euclidean distance between joint positions. Optimizing angles reduces the position loss as well, but optimizing the latter directly achieves lower errors and faster convergence.
      \textbf{\subref{fig:angle_loss_gradient}} Exploding gradients with the angle loss.
      \textbf{\subref{fig:position_loss_gradient}} Stable gradients with the position loss. In that case, noise is solely due to SGD sampling.}
    \label{fig:angle_vs_pos}
\end{figure*}

Our first experiment compares loss functions. We condition the generator on $n = 60$ frames and predict the next $k = 30$ frames. Figure~\ref{fig:angle_vs_pos}
shows that optimizing the angle loss can lead to larger position errors since it fails to properly assign credit to correct predictions on crucial joints. The angle loss is also prone to exploding gradients. 
This suggests that optimizing the position loss may reduce the complexity of the problem, which seems counterintuitive considering the overhead of computing forward kinematics. One possible explanation is that some postures may be difficult to optimize with angles, but if we consider motion as a whole, the model trained on position loss would make occasional mistakes on rotations without visibly affecting the result. Therefore, our forward kinematics positional loss is more attractive for minimizing position errors. Since this metric better reflects the quality of generation for long-term generation \citep{holden:deeplearning:2016}, we perform subsequent experiments with the position loss.

\begin{figure}
    \centering
      \subtable[]{
      \raisebox{.7\height}{
      \begin{adjustbox}{width=0.4\linewidth}
        \begin{tabular}{@{} c|c|c@{}}
          \multicolumn{3}{c}{Preference (\%)}\\
          Ours & None & Theirs\\\midrule
          41.4\% & 15.0\%  & 43.6 \%\\
          \bottomrule
          \noalign{\vskip 7mm}   
        \end{tabular}
        \end{adjustbox}
      	\label{tbl:mturk}
      	}
    }
    \subfigure[]{
        \includegraphics[height=0.4\linewidth]{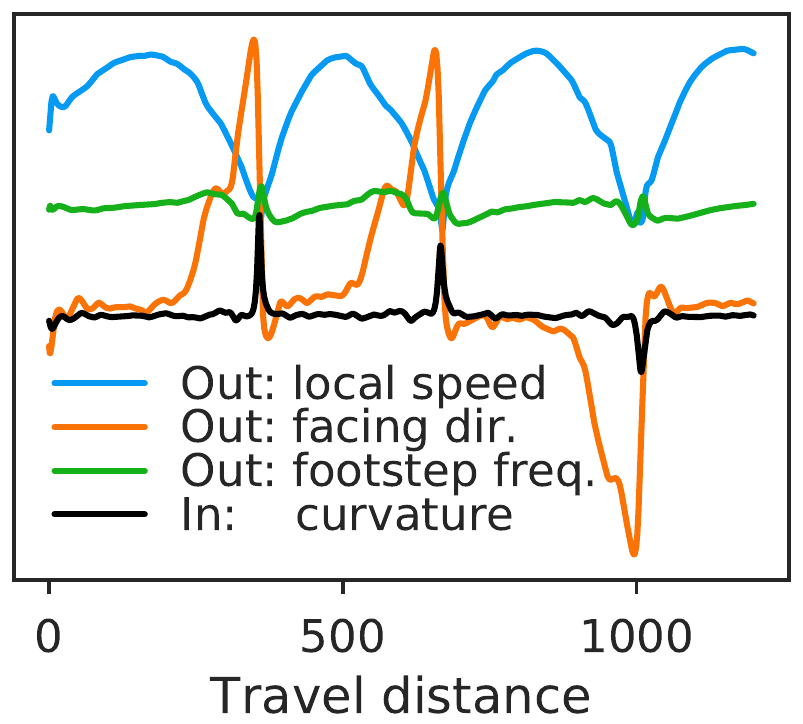}
        \label{fig:regressor_curves}
    }
    \caption{
    \textbf{\subref{tbl:mturk}} Human study comparing to \cite{holden:phase:2017}.
    \textbf{\subref{fig:regressor_curves}} Our \emph{pace network} allows fine control in space and time. Here, we instruct the character to sprint along a trajectory with sharp turns, represented as curvature spikes. The character anticipates turns by slowing down, rotating its body, and increasing the frequency of footsteps.}
\end{figure}

The second experiment assesses generation quality in a human study. We perform a side-by-side comparison with phase-functioned neural network \citep{holden:phase:2017}. For both methods, we generate 8 short clips ($\sim 15$ seconds) for walking along the same trajectory and for each clip, we collect judgments from 20 assessors hired through Amazon Mechanical Turk. We selected only workers with ``master'' status. Each task compared $5$ pairs of clips where methods are randomly ordered. Each task contains a control pair with an obvious flaw to exclude unreliable workers. Figure~\ref{tbl:mturk} shows that our method performs similarly to \cite{holden:phase:2017}, but without employing any post-processing.

\begin{figure*}[t]
\centering
\includegraphics[width=\textwidth]{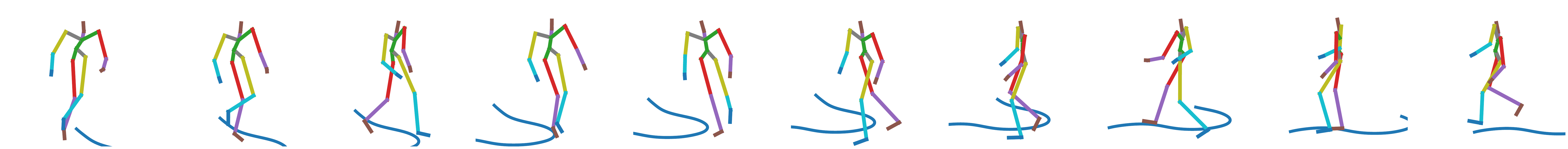}
\includegraphics[width=\textwidth]{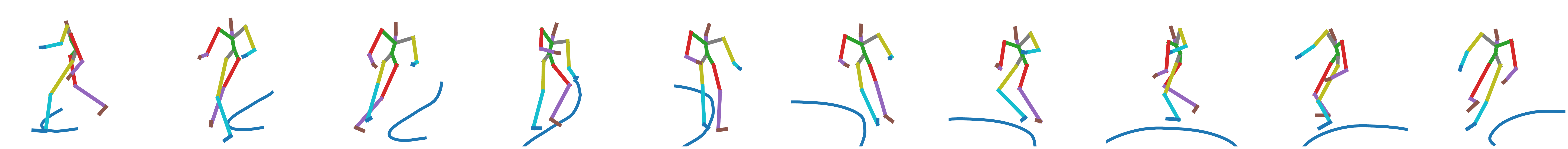}
\caption{Example of locomotion generation. \textbf{Above:} walking. \textbf{Below:} running.}
\label{fig:example}
\end{figure*}

Figure~\ref{fig:example} shows an example of our generation where the character is instructed to walk or run along a trajectory. Figure~\ref{fig:regressor_curves} shows how our \emph{pace network} computes the trajectory parameters given its curvature and a target speed. Our generation, while being online, follows exactly the given trajectory and allows for fine control of the time of passage at given way points. \citet{holden:deeplearning:2016} presents the same advantages, although these constraints are imposed as an offline post-processing step, whereas \citet{holden:phase:2017} is online but does not support time or space constraints.

\subsection{Ablations}
\label{sec:ablations}

In this section we compare different human pose representations and then ablate various hyperparameters to better understand the behavior of our model.

\subsubsection{Conditioning length}

First, we measure the effect of differently sized conditioning sequences $n$ (cf. Section~\ref{sec:method}). 
For the RNN model, we try $n = 1, 2, 5, 10, 25, 50$ and for the CNN model $n = 1, 2, 4, 8, 16, 32, 48$. 
For the CNN, $n$ corresponds to the size of the receptive field.

\begin{figure}
    \centering
    \subfigure[RNN\hspace{-7mm}]{
    	\includegraphics[width=0.47\linewidth]{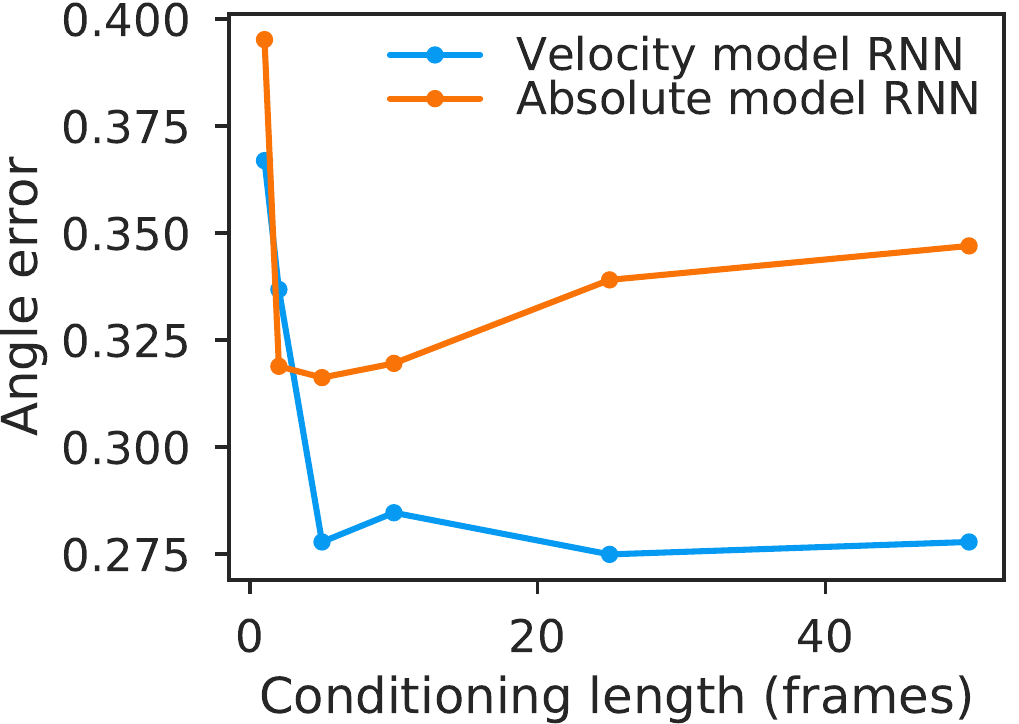}
		\label{fig:field_rnn}
    }
    \subfigure[CNN\hspace{-7mm}]{
    	\includegraphics[width=0.46\linewidth]{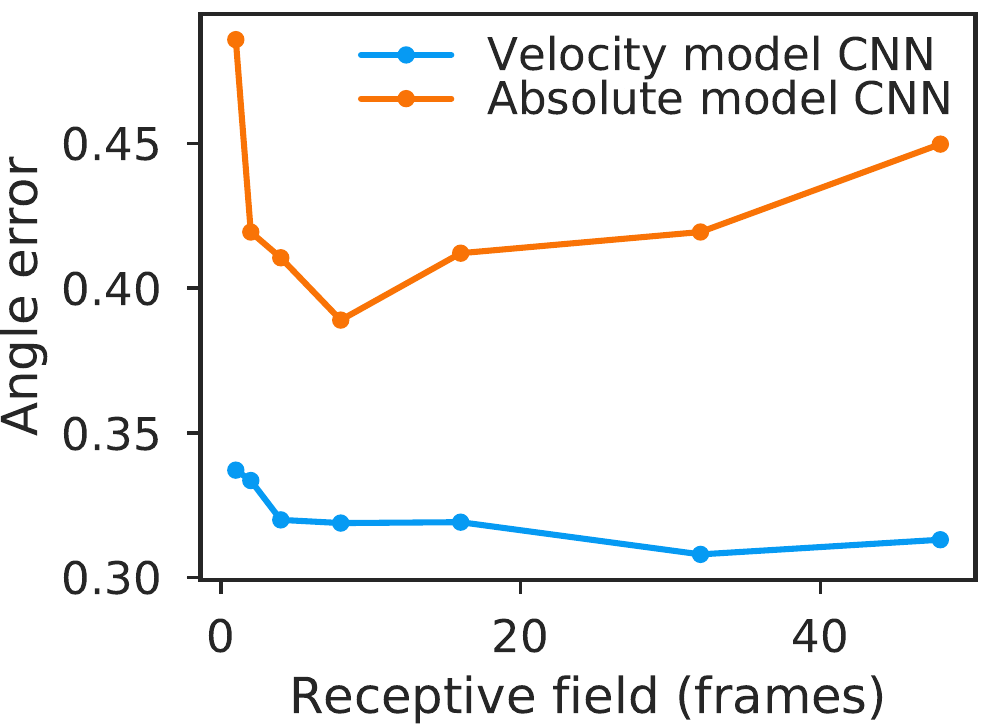}
		\label{fig:field_cnn}
    }
    \caption{Error as a function of the conditioning sequence length for RNN \subref{fig:field_rnn} and CNN \subref{fig:field_cnn} architectures and their respective velocity variations. 
    We show the angle error after 80 ms for action ``Walking''.}
    \label{fig:field}
\end{figure}

Figure~\ref{fig:field} shows that the error saturates after 10--20 frames (400--800 msec) for both models which is likely  because the models are not exploiting long-term information.
This is certainly in part due to the high level of uncertainty in predicting human motion: very old frames provide little information about the future since there are many possible predictions.
For the CNN with absolute rotations, large receptive fields are not necessarily best and smaller sizes often perform better.

\subsubsection{Parameterizations}

Next, we compare quaternions, Euler angles, and axis-angle vectors to parameterize rotations in the long-term generation setting (Section~\ref{sec:longterm} and Section~\ref{sec:longterm_exp}).
In addition to the position error, we also measure the \emph{velocity error}, defined as the Euclidean error of the first derivative of the position. 
The velocity loss is a good indicator of the smoothness of the generated poses. 
High velocity error is most likely due to jitter or discontinuities.
In order to compose rotations, we convert the output rotations to quaternions before feeding them to the forward kinematics layer.

\begin{figure}
    \centering
    \subfigure[Position loss\hspace{-5mm}]{
    	\includegraphics[width=0.48\linewidth]{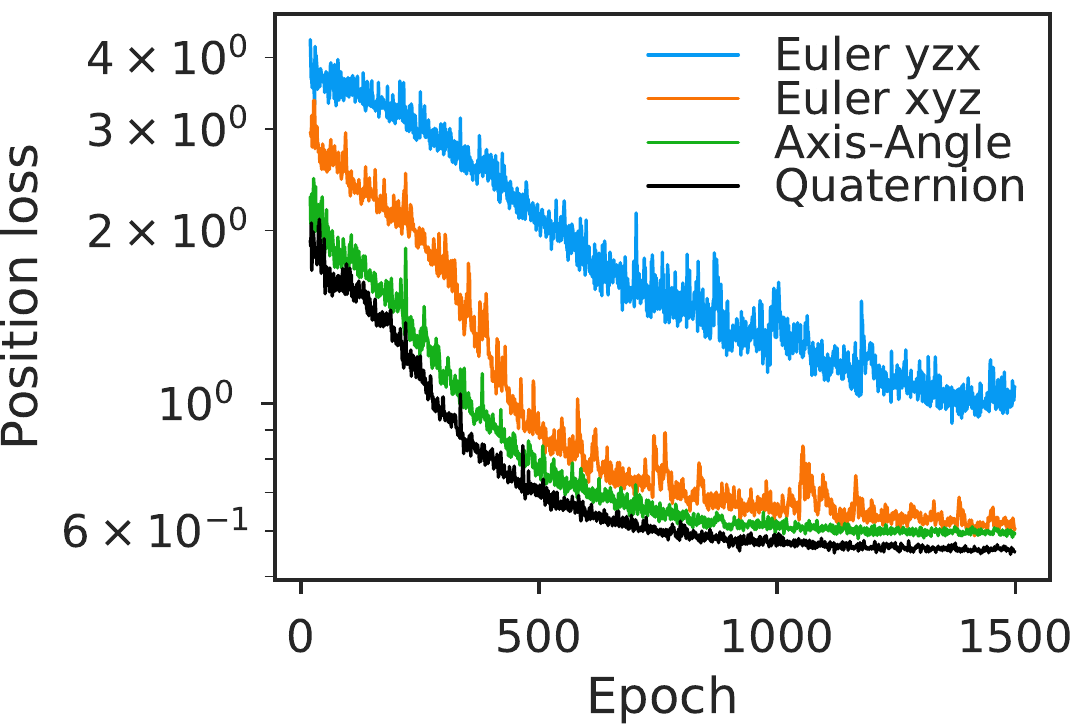}
		\label{fig:repr_mpjpe}
    }
    \subfigure[Velocity loss\hspace{-5mm}]{
    	\includegraphics[width=0.45\linewidth]{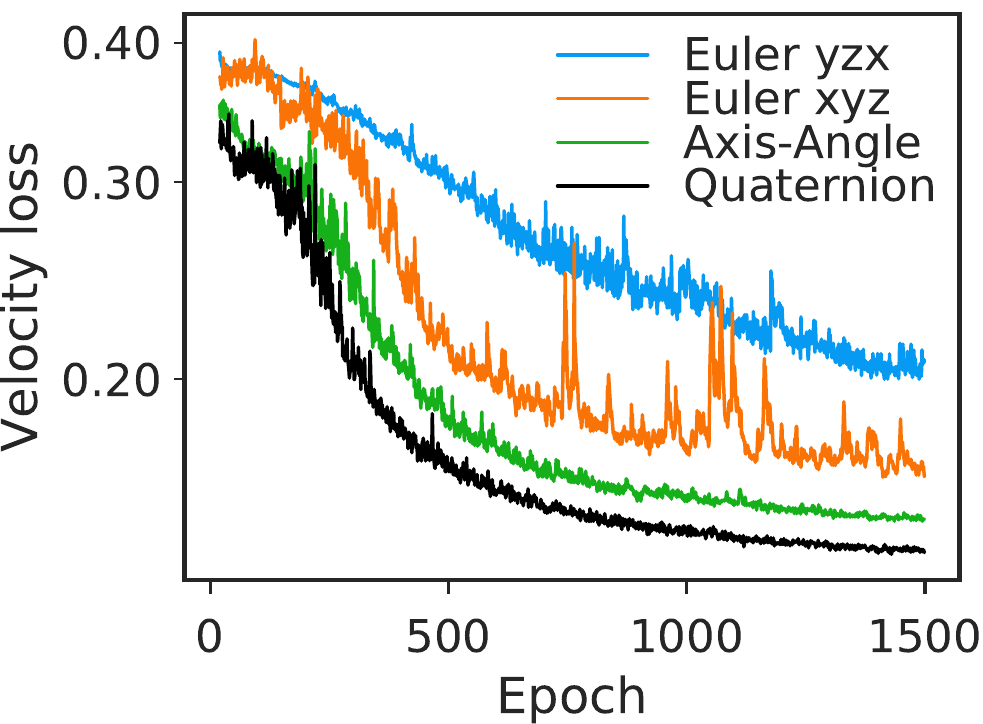}
		\label{fig:repr_mpjve}
    }
    \caption{Validation loss for different rotation parameterizations during training on the long-term task. 
    Quaternions converge faster and to a lower loss.}
    \label{fig:repr}
\end{figure}

The results (Figure~\ref{fig:repr}) show that quaternions have the lowest error as well as the fastest convergence rate. 
In terms of the position error, the difference between the quaternion and axis-angle representations is narrow, however, the velocity loss shows  that quaternions produce smoother predictions. 

Interestingly, the performance of Euler angles depends on the chosen order convention: the $yzx$ order results in many discontinuities and poor performance, whereas the $xyz$ order is close to the axis-angle performance on this dataset, arguably because it reflects the degrees of freedom of the skeleton. 
Nonetheless, the velocity error and at the error distribution (Figure~\ref{fig:eul_vel_hist}) indicate that Euler angles give rise to spurious discontinuities in the generated poses, which are undesirable from a qualitative perspective. 

\begin{figure}
    \centering
    \subfigure[Position loss\hspace{-5mm}]{
    	\includegraphics[width=0.45\linewidth]{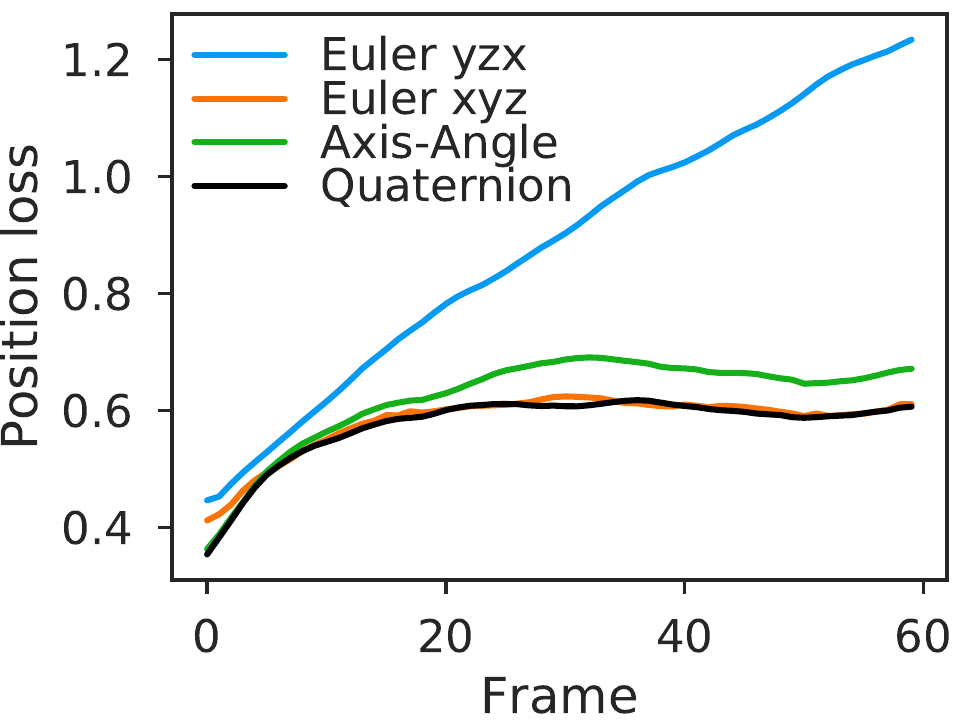}
		\label{fig:repr_frame_mpjpe}
    }
    \subfigure[Velocity loss\hspace{-5mm}]{
    	\includegraphics[width=0.47\linewidth]{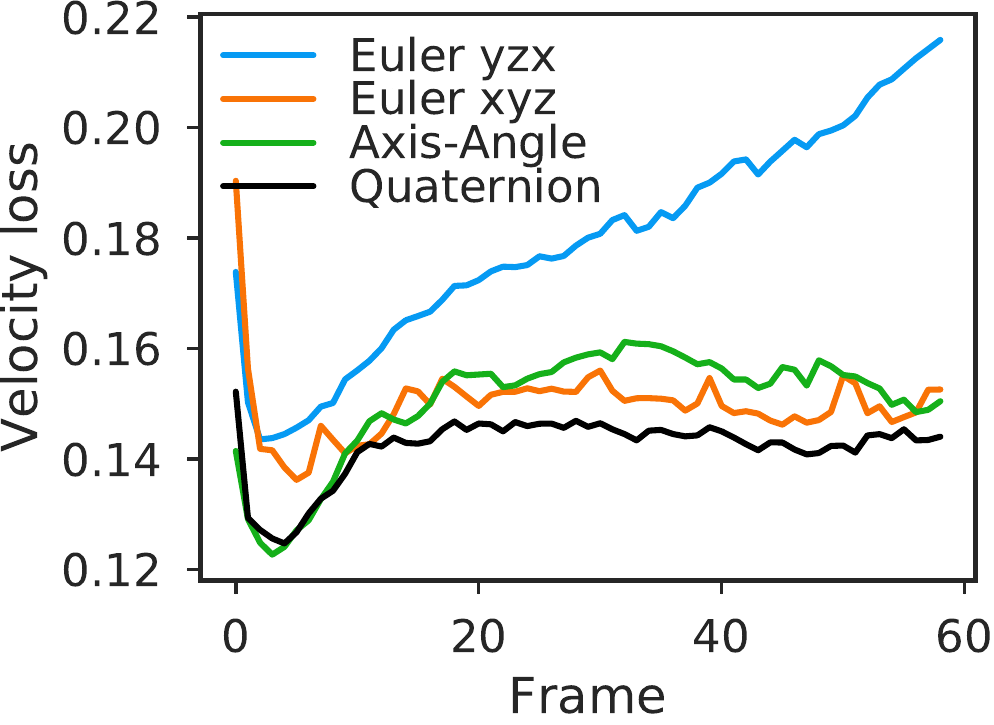}
		\label{fig:repr_frame_mpjve}
    }
    \caption{Comparison between rotation parameterizations. We show the error during inference on the long-term task, at different time horizons.}
    \label{fig:inference_error}
\end{figure}

Figure~\ref{fig:inference_error} shows inference time errors for predicting up to 60 frames into the future after models are fully trained.
The error quickly plateaus for quaternions but not so for axis-angle rotations and $yzx$ Euler angles. 
As before, $xyz$ Euler angles perform similarly to quaternions with respect to the position error but they perform less well in terms of the velocity error.

\subsubsection{Rotation vs position regression}

Generating joint rotations is required for some applications, e.g. for the animation of skinned meshes, and we can directly train a model to perform this task (Section~\ref{sec:rotations_vs_positions}).
An alternative is to predict 3D joint positions and to recover the joint rotations via inverse kinematics, implemented as a non-differentiable post-processing step~\citep{holden:phase:2017}. 
We compare the two approaches by comparing quaternion to a model that predicts joint positions (Position).
For the latter, we also consider projecting poses onto a valid skeleton by performing inverse kinematics followed by forward kinematics (Pos. reproj.). 
Specifically, we solve with projected gradient descent using the Adam optimizer \citep{kingma:adam:2014}, until convergence of the Euclidean error loss. 
In practice, many solvers use heuristics or converge to a suboptimal solution for performance reasons, but the goal of our experiment is to illustrate what lower bound can be achieved. 

\begin{figure}
    \centering
    \subfigure[Position\hspace{-4mm}]{
    	\includegraphics[width=0.45\linewidth]{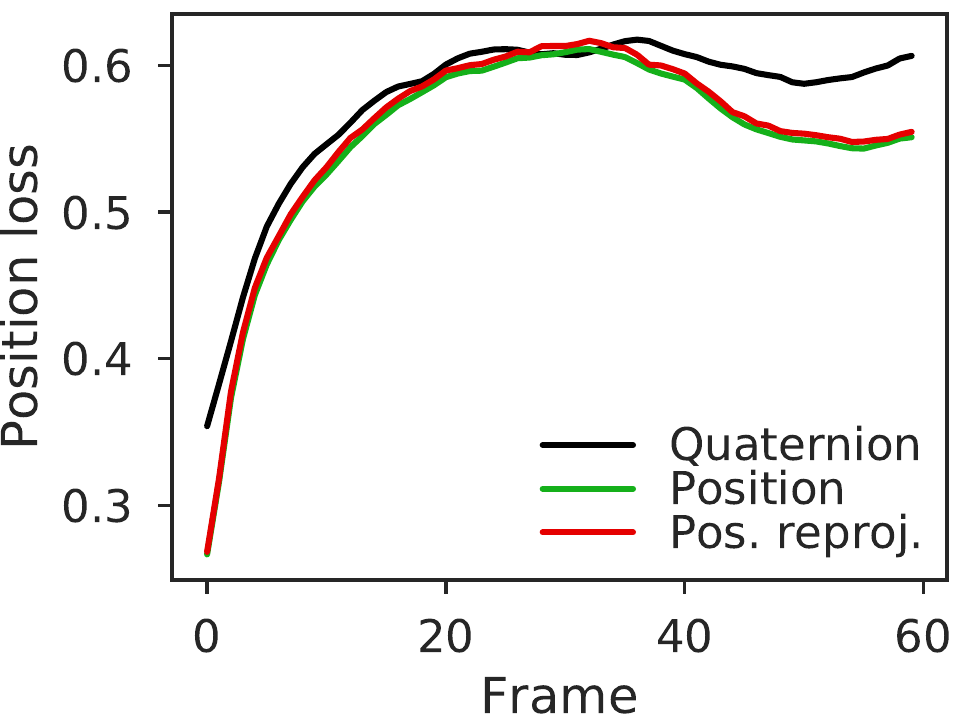}
		\label{fig:ik_pos}
    }
    \subfigure[Velocity\hspace{-5mm}]{
    	\includegraphics[width=0.47\linewidth]{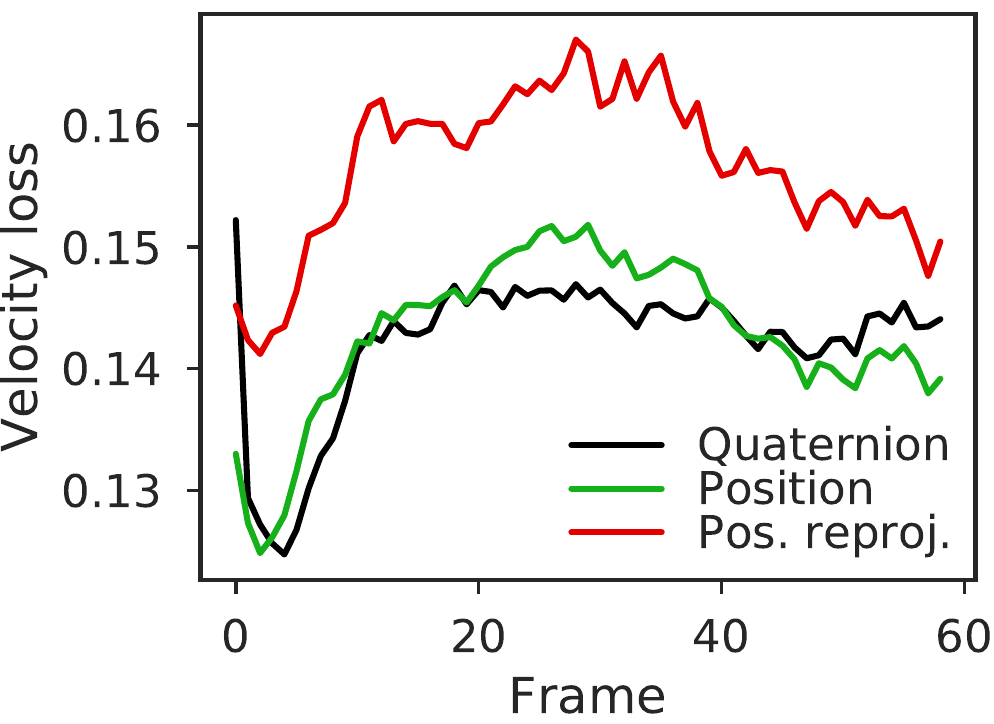}
		\label{fig:ik_vel}
    }
    \caption{Comparison between a model that outputs quaternions and one that outputs 3D joint positions (Position), on the long-term task. We also include the error after reprojecting the 3D pose onto a valid skeleton (Pos. reproj.).}
    \label{fig:ik}
\end{figure}

\begin{figure}
    \centering
    \subfigure[\hspace{-7mm}]{
    	\includegraphics[width=0.46\linewidth]{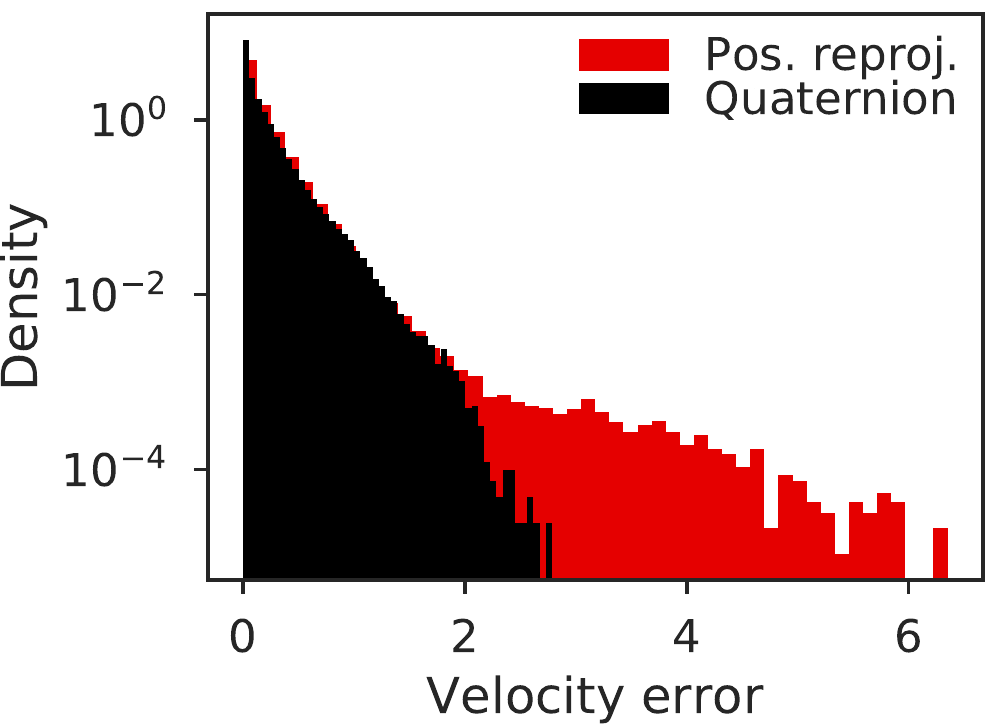}
		\label{fig:ik_vel_hist}
    }
    \subfigure[\hspace{-7mm}]{
    	\includegraphics[width=0.46\linewidth]{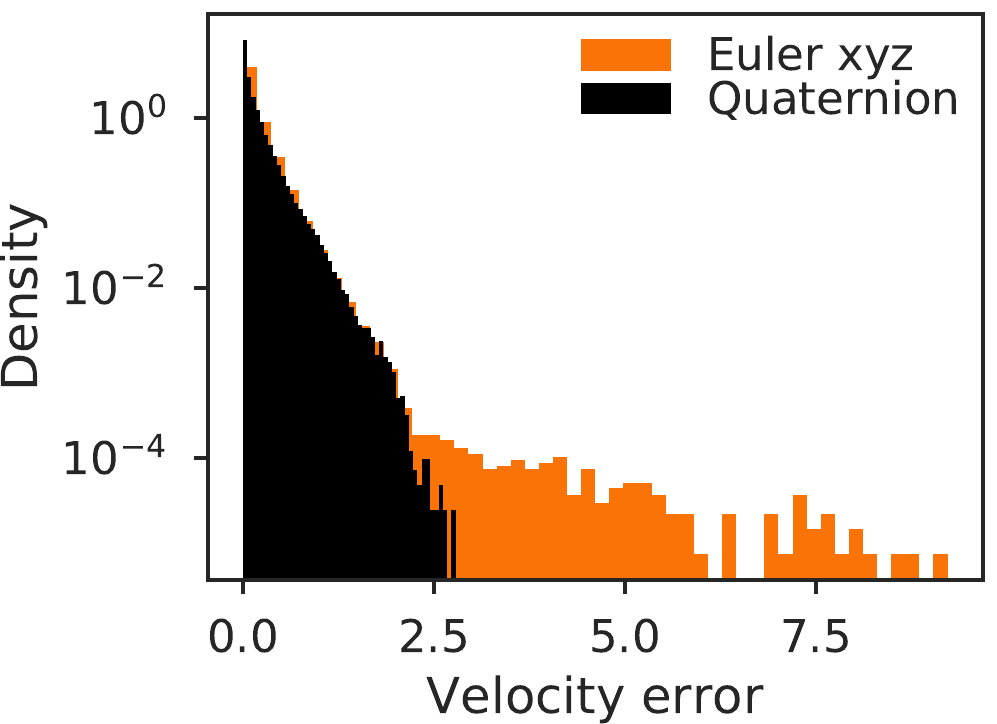}
		\label{fig:eul_vel_hist}
    }
    \caption{Velocity error histogram on the long-term task. \textbf{\subref{fig:ik_vel_hist}}~Reprojecting 3D poses via inverse kinematics introduces high-frequency jitter (Section~\ref{sec:rotations_vs_positions}). \textbf{\subref{fig:eul_vel_hist}}~Euler angles introduce high-frequency artifacts caused by the discontinuous representation space.}
\end{figure}

Figure~\ref{fig:ik} shows that all approaches achieve similar position loss. 
The quaternion model is slightly worse after 40 frames, most likely because of the higher complexity of the loss function. 
On the other hand, the velocity error after re-projection is higher than the quaternion model.
This is likely because position re-projection introduces discontinuities as illustrated in Figure~\ref{fig:ik_vel_hist}. In principle, it is possible to introduce a smoothness constraint in the solver, but this would further limit online processing.
Considering the computational cost of inverse kinematics and the lack of practical advantages, we argue that a model trained to predict joint rotations is more versatile.

\section{Conclusion and future work}
\label{sec:ccl}

We propose \name, a neural network architecture based on quaternions for rotation parameterization -- an overlooked aspect in previous work. 
Our experiments show the advantage of our model for both short-term prediction and long-term generation, while previous work typically addresses each task separately. 
We also suggest training with a position loss that performs forward kinematics on a parameterized skeleton. 
This benefits both from a constrained skeleton (like previous work relying on angle loss) and from proper weighting across different joint prediction errors (like previous work relying on position loss).  
Our results improve short-term prediction over the popular Human3.6M dataset, while our long-term generation of locomotion qualitatively compares with recent work in computer graphics.
Furthermore, our generation is real-time and allows better control of time and space constraints. 
Finally, we showed that the standard evaluation protocol for the Human3.6M dataset produces high-variance results and we propose a simple solution.

As for future work, \name can be extended to tackle other motion-related tasks, such as action recognition or pose estimation from video. 
In this regard, a promising research direction is represented by \emph{self-supervised} pose estimation, which can benefit from a parameterized skeleton in the supervision signal. 
Another trend is \emph{weakly supervised} training, where one model generates training data for another model on a different task. 
For instance, it would be interesting to train \name on low-quality poses inferred from video. 
For motion generation, this would provide further artistic control with additional inputs and would enable conditioning based on a richer set of actions.

Another promising research direction is neural networks that perform computations directly in quaternionic domain. 
Currently, \name uses standard RNN and CNN architectures as its backbone which operate in Euclidean space. 
Recently, quaternion-valued RNNs~\citep{parcollet:quaternion:2018} and CNNs~\citep{zhu:quaternioncnn:2018, gaudet:deep:2018, parcollet:quaternion_speech:2018} have been proposed, resulting in promising results on tasks with long-range dependencies such as speech recognition. 
These architectures would be interesting for human motion modeling.

Orthogonal to our work is also the question of generative model training: we use step-wise regression and scheduled sampling~\citep{bengio:scheduled:2015}. 
Very recent work has shown state-of-the-art results with adversarial training that contrasts model samples with real data~\citep{gui:adversarial:2018}. 
Pairing adversarial training with quaternion-parameterized kinematics is an interesting future avenue.

\bibliographystyle{spbasic}
\bibliography{main}
\end{document}